\documentclass[10pt]{article} 
\usepackage[accepted]{tmlr}

\usepackage{amsmath,amsfonts,bm}

\def\eqref#1{equation~\ref{#1}}

\def\1{\bm{1}}

\DeclareMathAlphabet{\mathsfit}{\encodingdefault}{\sfdefault}{m}{sl}
\SetMathAlphabet{\mathsfit}{bold}{\encodingdefault}{\sfdefault}{bx}{n}

\usepackage{hyperref}
\definecolor{myblue}{HTML}{598BE7}
\hypersetup{
    colorlinks = true,
    citecolor  = myblue,
    linkcolor  = myblue,
    urlcolor   = myblue,
    pdfborder  = {0 0 0} 
}
\usepackage{url}

\usepackage{pgfplots}
\usepackage{tikz}
\usepackage{subcaption}
\usepackage{booktabs}
\usepackage{multirow}
\usepackage{pgfplotstable}
\usepgfplotslibrary{fillbetween}
\usepackage{anyfontsize}
\usepackage{tabularray}
\usepackage{tabularx}

\usepackage{array}
\usepackage{booktabs} 
\usepackage{pifont}
\usepackage{amsmath} 
\usepackage{amssymb}
\usepackage{wasysym}
\usepackage{algorithm}
\usepackage{algpseudocode}
\usepackage{adjustbox}
\renewcommand{\eqref}[1]{(\ref{#1})}

\newcommand{\sftype}[1]{{\textsf{\small #1}}}

\newcommand{\cmark}{\ding{51}}
\newcommand{\xmark}{\ding{55}}

\usepackage{color, colortbl}
\usepackage{float}

\definecolor{cBlue}{HTML}{1852CC}
\definecolor{cBlue2}{HTML}{3fb8b8}
\definecolor{cBlue3}{HTML}{02ffff}
\definecolor{cBlue4}{HTML}{00b5ff}
\definecolor{cRed}{HTML}{D62728}
\definecolor{cRed2}{HTML}{ED521F}
\definecolor{cRed3}{HTML}{F69C40}
\definecolor{cRed4}{HTML}{fa4d3a}
\definecolor{cGreen}{HTML}{2CA02C}
\definecolor{cGreen2}{HTML}{3fdf3f}
\definecolor{cPink}{HTML}{ED1FD2}
\definecolor{cWhite}{HTML}{ffffff}
\definecolor{Purple}{HTML}{b05cff}
\definecolor{Gray}{gray}{0.9}
\definecolor{Salmon}{HTML}{FF7E79}
\definecolor{Orchid}{HTML}{7A81FF}
\definecolor{cBlue5}{HTML}{409BA0}
\definecolor{cPink2}{HTML}{CB2CED}
\definecolor{cPink3}{HTML}{ED1D81}
\definecolor{PastelPink}{HTML}{FC94AF}

\newcommand{\khy}[1]{{\color{cBlue}{#1}}}

\usetikzlibrary{plotmarks}
\usepackage{xspace} 
\providecommand{\ie}{\textit{i.e.}\@\xspace}

\title{Active Prompt Learning with Vision-Language Model Priors}

\author{\name Hoyoung Kim
\email hoyoung.kim@postech.ac.kr \\
\addr Graduate School of Artificial Intelligence \\ POSTECH
\AND
\name Seokhee Jin
\email jin749@postech.ac.kr \\
\addr Graduate School of Artificial Intelligence \\ POSTECH
\AND
\name Changhwan Sung
\email changhwan.sung@postech.ac.kr \\
\addr Graduate School of Artificial Intelligence \\ POSTECH
\AND
\name Jaechang Kim
\email jaechang@postech.ac.kr \\
\addr Graduate School of Artificial Intelligence \\ POSTECH
\AND
\name Jungseul Ok
\email jungseul.ok@postech.ac.kr \\
\addr Graduate School of Artificial Intelligence \\ POSTECH
}

\def\month{10}  
\def\year{2025} 
\def\openreview{\url{https://openreview.net/forum?id=qBeGCzD3Ij}} 

\begin{document}

\maketitle

\begin{abstract}
Vision-language models (VLMs) have demonstrated remarkable zero-shot performance across various classification tasks. Nonetheless, their reliance on hand-crafted text prompts for each task hinders efficient adaptation to new tasks. While prompt learning offers a promising solution, most studies focus on maximizing the utilization of given few-shot labeled datasets, often overlooking the potential of careful data selection strategies, which enable higher accuracy with fewer labeled data. This motivates us to study a budget-efficient active prompt learning framework. Specifically, we introduce a class-guided clustering that leverages the pre-trained image and text encoders of VLMs, thereby enabling our cluster-balanced acquisition function from the initial round of active learning. Furthermore, considering the substantial class-wise variance in confidence exhibited by VLMs, we propose a budget-saving selective querying based on adaptive class-wise thresholds. Extensive experiments in active learning scenarios across seven datasets demonstrate that our method outperforms existing baselines.
\end{abstract}
\section{Introduction}
\label{sec:intro}
Vision-language models (VLMs), such as CLIP~\citep{radford2021learning} and ALIGN~\citep{jia2021scaling}, have demonstrated impressive zero-shot capabilities across various downstream tasks, including object detection~\citep{du2022learning,feng2022promptdet,zhong2022regionclip}, semantic segmentation~\citep{yi2023simple,ghiasi2022scaling,lilanguage}, and image classification~\citep{radford2021learning,singh2022flava,zhai2022lit}, by aligning visual and textual information within a shared representation space~\citep{radford2021learning,jia2021scaling,yuan2021florence}.
Nevertheless, as VLMs rely on manually crafted text prompts for each task, which can be time-consuming and labor-intensive, efficiently adapting VLMs to new tasks remains crucial.
Prompt learning has emerged as a promising solution, particularly for image classification tasks, allowing VLMs to learn task-specific prompts without the computational burden of directly fine-tuning image and text encoders.

Generally, prompt learning methods have focused on model-centric approaches, modifying prompt architectures and learning objectives.
Specifically, researchers have introduced various prompt types, including text prompts~\citep{zhou2022learning}, image-conditioned text prompts~\citep{zhou2022conditional} for text encoders, and multimodal prompts that work across both image and text encoders~\citep{khattak2023maple}.
In terms of learning objectives, prompts are initially trained with cross-entropy loss~\citep{zhou2022learning}, supplemented by regularization terms to maintain CLIP's general knowledge~\citep{zhu2023prompt} and incorporate task-agnostic knowledge~\citep{park2024prompt}, which helps prevent overfitting on specific tasks.
However, these model-centric methods mainly focus on leveraging VLM priors to optimize prompts on given few-shot labeled datasets, often overlooking the potential of data selection to achieve higher accuracy with fewer labeled samples.

In contrast to previous model-centric approaches~\citep{zhou2022learning,zhu2023prompt}, we adopt a data-centric perspective by explicitly leveraging VLM priors to select informative data.
In this context, active learning offers an alternative by prioritizing the labeling of the most informative images with minimal budgets.
Recent work on active prompt learning~\citep{bang2024active} highlights that class-balanced data selection is crucial for mitigating the inherent imbalanced knowledge within VLMs.
However, this approach merely adheres to the conventional few-shot datasets, disregarding the opportunity to fully utilize VLM priors. 
In contrast, we fully exploit VLM priors through our proposed class-guided clustering and selective querying methods.

\begin{figure*}[t!]
    \centering
    \includegraphics[width=\textwidth]{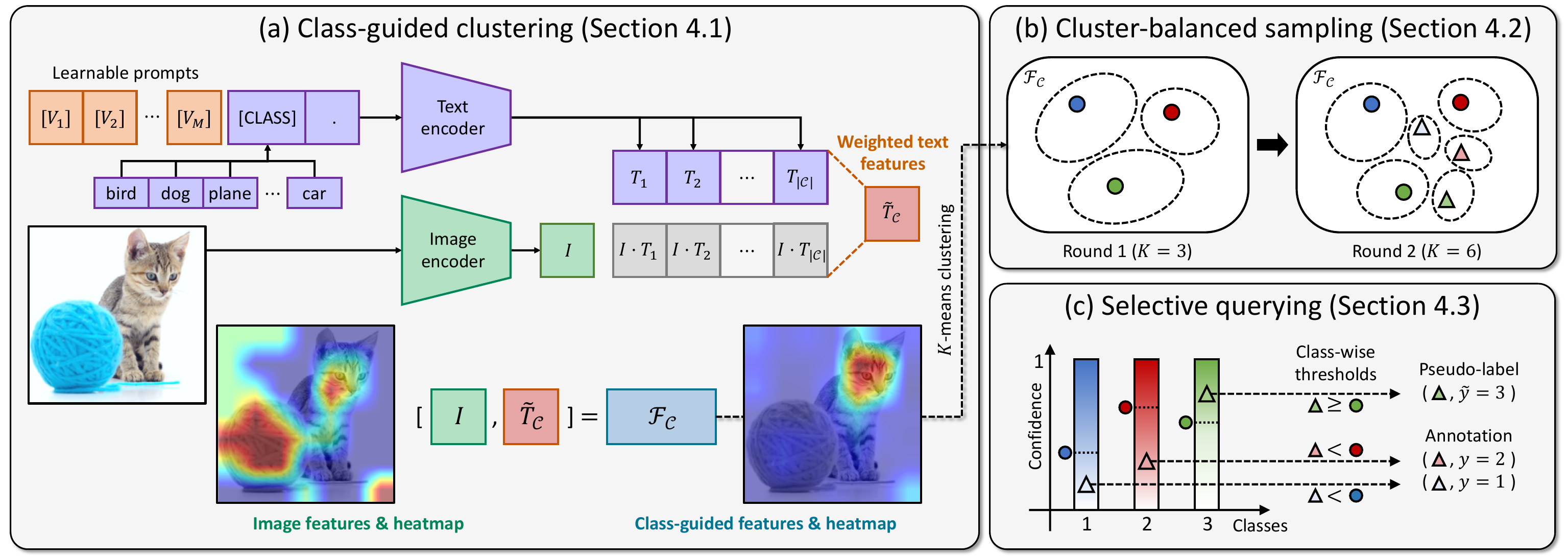}
    \caption{{\em An overview of the proposed framework.} (a) Class-guided features $\mathcal{F}_{\mathcal{C}}$ are obtained by concatenating the image features $I$ with the weighted text features ${\tilde T}_{\mathcal{C}}$, using similarity scores as weights. In the heatmaps, $\mathcal{F}_{\mathcal{C}}$ focus on the guided-classes $\mathcal{C}=\{\text{Cat},\text{Dog}\}$ than $I$. 
    (b) $K$-means clustering is performed on $\mathcal{F}_{\mathcal{C}}$. With an increasing $K$, cluster-balanced sampling becomes available in each round. 
    (c) The confidence scores of previously labeled data (circles) serve as thresholds for new candidates (triangles). If a candidate's confidence exceeds its corresponding threshold, a pseudo-label is assigned to conserve the budget, otherwise it is labeled by annotators.}
    \vspace{-3mm}
    \label{fig:main_method}
\end{figure*}

For a class-guided clustering, we leverage the pre-trained image and text encoder of VLMs.
In detail, we first derive class-guided features by concatenating two components: (i) image features from the image encoder and (ii) text features computed as a weighted sum of each class's text features, with weights based on similarity scores to the image features.
We then apply $K$-means clustering~\citep{macqueen1967some} on these class-guided features to achieve balanced data selection across clusters.
While traditional active learning often encounters the cold-start problem due to a lack of reliable methods to evaluate data in the initial round~\citep{mahmood2021low,chen2023making}, our cluster-balanced acquisition function provide the benefit of a warm-start.
However, purely diversity-focused acquisitions may inefficiently allocate budgets to data in which VLMs are already confident.

To address this issue, we introduce a budget-saving selective querying based on adaptive class-wise thresholds.
Since VLMs often exhibit substantial variance in confidence across different downstream tasks and even among individual classes~\citep{bang2024active}, we implement adaptive class-wise thresholds without adding extra hyperparameters.
Specifically, we assign pseudo-labels to the data selected by the class-balanced acquisition when their confidence scores exceed the corresponding threshold.
As a result, we can conserve budget in each round rather than exhausting it entirely.


The proposed framework, illustrated in Figure~\ref{fig:main_method}, fully leverages VLM priors to enable efficient adaptation across various classification tasks.
Beyond simply employing the two encoders of CLIP~\citep{radford2021learning} for clustering~\citep{liimage}, we further analyze the advantages of incorporating the weighted text features through visualization tools, such as GradCAM~\citep{selvaraju2017gradcam} in Figure~\ref{fig:heatmap} and T-SNE~\citep{van2008visualizing} in Figure~\ref{fig:tsne-birds}.
Extensive experiments in active learning scenarios demonstrate that our cluster-balanced acquisition with selective querying outperforms other baselines.

Our main contributions are summarized as follows:
\begin{itemize}
    \vspace{-1.5mm}
    \item We propose a budget-efficient active prompt learning for VLMs, particularly on CLIP, where the class-guided clustering and the selective querying fully leverage VLM priors (Sections~\ref{sec:multimodal-clustering} and~\ref{sec:budget-saving}).
    \item We provide in-depth analyses of the class-guided features and clustering with GradFAM, a variant of GradCAM, and T-SNE, respectively (Figures~\ref{fig:heatmap} and~\ref{fig:tsne-birds}).
    \item Experiments demonstrate that our method achieve superior budget efficiency and performance across diverse active learning scenarios (Sections~\ref{sec:active-learning-scenario}).
    \item We explore the potential for extending of our data-centric approach into existing model-centric prompt learning methods (Section~\ref{sec:further-analyses}).
\end{itemize}

\section{Related Work}
\textbf{Prompt learning in vision-language models.}
To address the inefficiency of fine-tuning all VLM parameters, CoOp~\citep{zhou2022learning} proposes prompt learning focused on compact prompts for efficient adaptation.
Subsequent works~\citep{zhou2022conditional,khattak2023maple, khattak2023self,park2024prompt,li2023gradient} have further developed CoOp, adopting model-centric approaches such as modifying prompt architectures~\citep{khattak2023maple} and learning objectives~\citep{zhu2023prompt}.
For instance, MaPle~\citep{khattak2023maple} incorporates multi-modal prompts that jointly consider both VLM encoders, while ProGrad~\citep{zhu2023prompt} introduces an auxiliary loss term to maintain general knowledge of CLIP. In contrast, PCB~\citep{bang2024active} efficiently adapts VLMs from a data-centric perspective by employing an active learning with a pseudo class-balanced acquisition function.
While PCB's acquisition focuses on mitigating the imbalanced prior knowledge of VLMs, we fully exploit VLM priors across the proposed method.

\vspace{1mm}
\textbf{Active learning in the era of foundation models.}
We are in an era where foundation models, such as CLIP~\citep{radford2021learning}, SAM~\citep{kirillov2023segment}, and GPT-4~\citep{achiam2023gpt}, dominate a wide range of downstream tasks.
Their impressive generalization capabilities may imply that the role of active learning is becoming less significant.
However, recent studies continue leveraging the prior knowledge embedded in foundation models~\citep{kimactive,gupte2024revisiting,bayer2024activellm,wan2023survey} to further enhance budget efficiency with active learning.
For instance, ALC~\citep{kimactive} introduces correction queries to refine SAM-generated labels in semantic segmentation tasks, while ActiveLLM~\citep{bayer2024activellm} addresses the cold start problem by utilizing LLMs in text classification tasks.
We fully leverage VLM priors to enhance budget efficiency in active learning for image classification tasks.


\vspace{1mm}
\textbf{Acquisition functions in active learning.}
Active learning employs various acquisition functions to identify the most informative samples for annotation, aiming to maximize model performance within a constrained budget.
These acquisitions are broadly categorized into those focused on uncertainty~\citep{asghar2016deep,he2019towards,ostapuk2019activelink,fuchsgruber2024uncertainty},
diversity~\citep{sener2018active,sinha2019variational,yehuda2022active}, and both~\citep{ash2019deep,hwang2022combating,Kim_2023_ICCV,wang2019incorporating,NEURIPS2023_559a0998}.
Recent studies show that uncertainty-based acquisitions are more effective with higher budgets, while diversity-based ones perform better with lower budgets~\citep{hacohen2022active,NEURIPS2023_2b09bb02,hacohen2023how}.
Building on this, we propose a cluster-balanced acquisition to adapt VLMs within limited budgets.
In addition, we introduce a selective querying with adaptive class-wise thresholds to further conserve budget.

\section{Preliminaries}
For efficient adaptation in vision-language models (VLMs), we fully leverage their prior knowledge.
Before presenting our method, we first outline the structure of VLMs and the relevant priors, in Section~\ref{sec:prior}, which are utilized in our method.
After that, we describe the basics of prompt learning with few-shot datasets in Section~\ref{sec:prompt-learning}.

\subsection{VLM Priors}
\label{sec:prior}
Pre-trained VLMs have shown decent zero-shot performance across various classification tasks.
Specifically, the CLIP model $\theta$~\citep{radford2021learning} comprises an image encoder $\theta_{\text{img}}$ and a text encoder $\theta_{\text{txt}}$, performing zero-shot inference on an image $x$ for a target class $c \in \mathcal{C}$ using cosine similarity as:
\begin{align}
\label{eq:prob-vec}
p_\theta(y = c \mid x; t,\mathcal{C}) := \frac{\text{exp} \big( \text{cos} (\theta_{\text{img}}(x) , \theta_{\text{txt}}(t_c)) / \tau \big)}{\sum_{k \in \mathcal{C}} \text{exp} \big( \text{cos} (\theta_{\text{img}}(x) , \theta_{\text{txt}}(t_k)) / \tau \big)} \;,
\end{align}
where the text prompt $t$ can be set as “a photo of a ”, and $t_c$ represents the concatenation of $t$ and “[CLASS].” with the class token replaced by the class name $c$.
Here, $\tau$ denotes a temperature parameter.
Considering the highest probability among all classes, we can obtain the pseudo label of an image $x$ as follows:
\begin{align}
\label{eq:pseudo-label}
y_\theta (x ; t, \mathcal{C}) := \arg \max_{c \in \mathcal{C}} p_\theta(y = c \mid x; t, \mathcal{C}) \;.
\end{align}
We note that the pseudo label depends on the text prompt $t$ and the target class set $\mathcal{C}$.
In Section~\ref{sec:method}, we leverage the pre-trained image and text encoders of CLIP for our class-guided clustering and the pseudo label for our selective querying.

\begin{algorithm}[t!]
\caption{Proposed Active Prompt Learning}
\begin{algorithmic}[1]
\Require
Image set $\mathcal{I}$, initial prompts $t_0$, budget per round $B$, and the number of rounds~$R$
\For{$r = 1, 2, \dots, R$}
    \State Extract class-guided features $\forall i \in \mathcal{I}$ by combining image and weighted text features via~\eqref{eq:multimodal-embeddings} 
    \State Perform K-means clustering on class-guided features 
    \State Select representative images from each cluster as candidates via~\eqref{eq:closest}
    \State Compute class-wise confidence thresholds using previously labeled data via~\eqref{eq:thresholds}
    \State Construct dataset $\mathcal{D}_r$ by assigning pseudo or ground-truth labels via selective querying in~\eqref{eq:dataset}
    \State Reinitialize and train prompts $t_r$ using dataset $\mathcal{D}_r$ with the objective in~\eqref{eq:final-loss}
\EndFor
\State \textbf{return} Final dataset $\mathcal{D}_R$ and trained prompts $t_R$
\end{algorithmic}
\label{alg1}
\end{algorithm}

\subsection{Prompt Learning in VLMs}
\label{sec:prompt-learning}
Vision-language models (VLMs) contain numerous parameters, which makes fine-tuning on a small labeled dataset impractical. 
Recently, prompt learning has facilitated efficient adaptation in VLMs by freezing the image and text encoders and focusing on learning input prompts.
For instance, CoOp~\citep{zhou2022learning} introduces learnable vectors into the text prompt $t_c$ of class $c$, replacing the conventional “a photo of a [CLASS].” text prompts with:
\begin{align}
t_c := [V_1] [V_2] \ldots [V_M] [\text{CLASS}]. \;,
\label{eq:learnable-vectors}
\end{align}
where each $V$ represents a learnable vector and $M$ denotes the number of vectors.
These vectors in $t$ are trained on a dataset $\mathcal{D}$ by minimizing the cross-entropy (CE) loss:
\begin{align}
\hat{\mathbb{E}}_{(x, y) \sim {\mathcal{D}}} \big[ \text{CE} \big( y, p_\theta (y; x, t, \mathcal{C}) \big) \big] \;.
\end{align}
In this context, previous prompt learning methods have primarily taken a model-centric perspective, aiming to maximize model performance on the given training dataset $\mathcal{D}$.
From a data-centric perspective, however, building datasets requires human labor, making active learning essential for creating information-dense datasets while minimizing interactions with annotators.
In addition, we note that our data-centric approach is compatible with existing model-centric prompt learning methods.

\section{Proposed Active Prompt Learning}
\label{sec:method}

\begin{figure*}[!t]
    \centering   
    \hspace{0.3mm}
    \begin{subfigure}[h!]{0.18\linewidth}
        \centering
        \setlength{\fboxsep}{0pt}\fbox{\includegraphics[width=1.0\linewidth, height=1.0\linewidth]{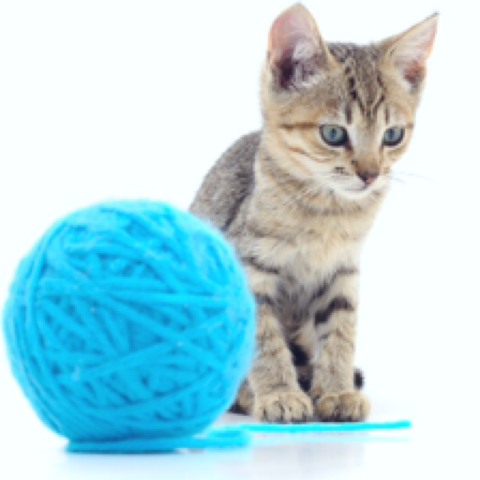}}
    \end{subfigure}
    \hspace{1mm}
    \begin{subfigure}[h!]{0.18\linewidth}
        \centering
        \setlength{\fboxsep}{0pt}\fbox{\includegraphics[width=1.0\linewidth, height=1.0\linewidth]{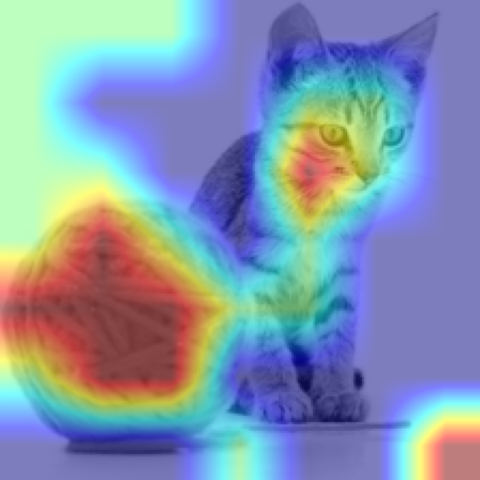}}
    \end{subfigure} %
    \hspace{1mm}
    \begin{subfigure}[h!]{0.18\linewidth}
        \centering
        \setlength{\fboxsep}{0pt}\fbox{\includegraphics[width=1.0\linewidth, height=1.0\linewidth]{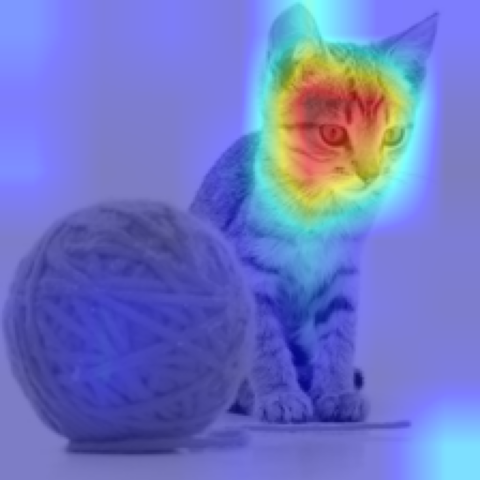}}
    \end{subfigure} %
    \hspace{1mm}
    \begin{subfigure}[h!]{0.18\linewidth}
        \centering
        \setlength{\fboxsep}{0pt}\fbox{\includegraphics[width=1.0\linewidth, height=1.0\linewidth]{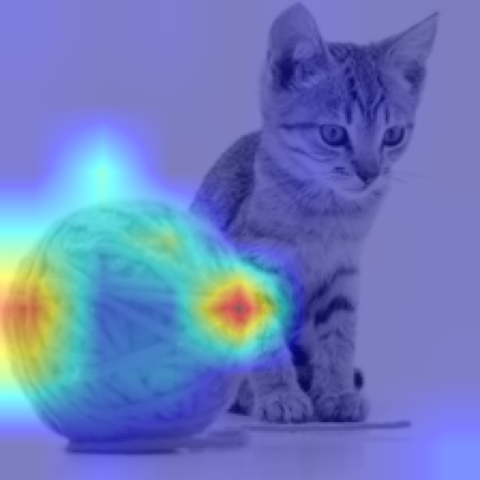}}
    \end{subfigure}
    \hspace{1mm}
    \begin{subfigure}[h!]{0.18\linewidth}
        \centering
        \setlength{\fboxsep}{0pt}\fbox{\includegraphics[width=1.0\linewidth, height=1.0\linewidth]{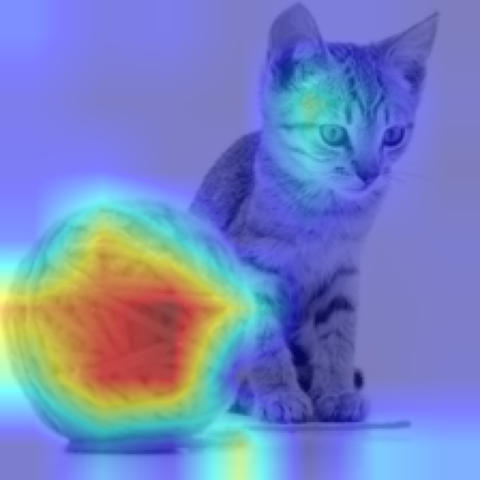}}
    \end{subfigure}
    \vspace{3mm}
    \\
    \centering
    \begin{subfigure}[h!]{0.18\linewidth}
        \centering
        \setlength{\fboxsep}{0pt}\fbox{\includegraphics[width=1.0\linewidth, height=1.0\linewidth]{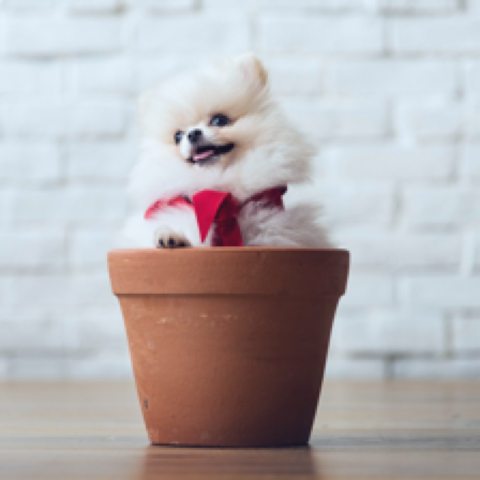}}
        \caption{Image}
        \label{subfig:heatmap-a}
    \end{subfigure}
    \hspace{1mm}
    \begin{subfigure}[h!]{0.18\linewidth}
        \centering
        \setlength{\fboxsep}{0pt}\fbox{\includegraphics[width=1.0\linewidth, height=1.0\linewidth]{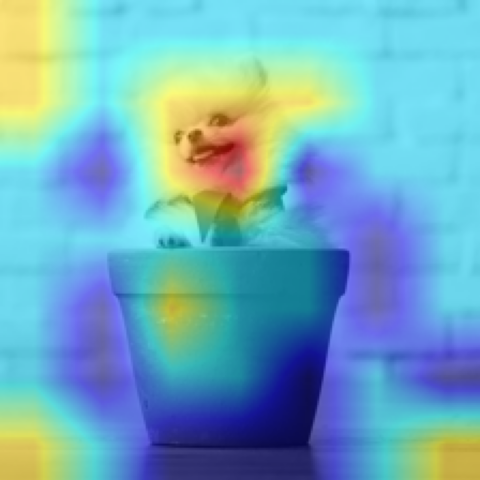}}
        \caption{$I$}
        \label{subfig:heatmap-b}
    \end{subfigure} %
    \hspace{1mm}
    \begin{subfigure}[h!]{0.18\linewidth}
        \centering
        \setlength{\fboxsep}{0pt}\fbox{\includegraphics[width=1.0\linewidth, height=1.0\linewidth]{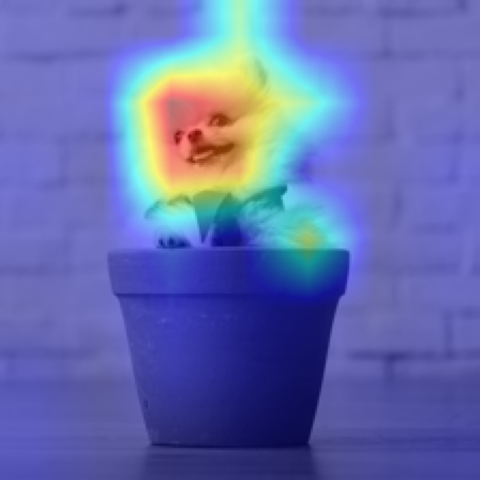}}
        \caption{$\mathcal{F}_{\mathcal{C}=\{\text{Cat}, \text{Dog}\}}$}
        \label{subfig:heatmap-c}
    \end{subfigure} %
    \hspace{1mm}
    \begin{subfigure}[h!]{0.18\linewidth}
        \centering
        \setlength{\fboxsep}{0pt}\fbox{\includegraphics[width=1.0\linewidth, height=1.0\linewidth]{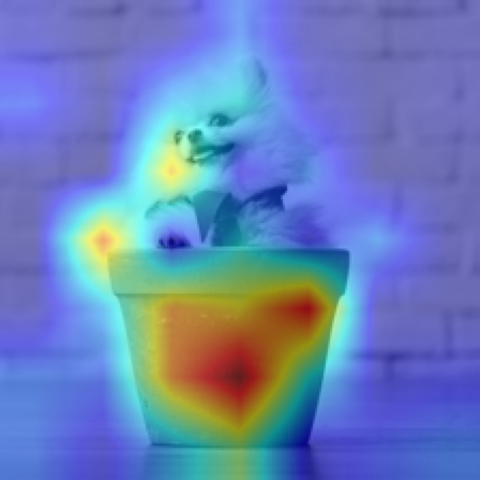}}
        \caption{$\mathcal{F}_{\mathcal{C}=\{\text{Ball}, \text{Pot}\}}$}
        \label{subfig:heatmap-d}
    \end{subfigure}
    \hspace{1mm}
    \begin{subfigure}[h!]{0.18\linewidth}
        \centering
        \setlength{\fboxsep}{0pt}\fbox{\includegraphics[width=1.0\linewidth, height=1.0\linewidth]{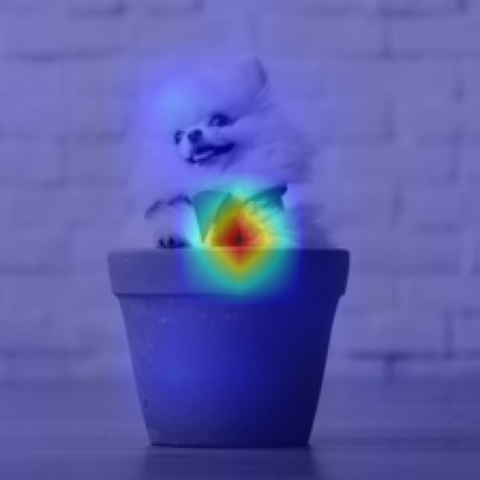}}
        \caption{$\mathcal{F}_{\mathcal{C}=\{\text{Red}, \text{Blue}\}}$}
        \label{subfig:heatmap-f}
    \end{subfigure}
    \caption{
        {\em GradFAM with various target features.} 
        (b) All objects in the image significantly impacts the target image features $I$. 
        (c-e) With our class-guided features $\mathcal{F}_{\mathcal{C}}$ for target features, the heatmap aligns with the target classes $\mathcal{C}$. Further details are in the Appendix~\ref{sec:heatmap_details}.
    }
    \label{fig:heatmap}
\end{figure*}

Given an unlabeled image set $\mathcal{U}$, we consider a general active learning scenario for image classification tasks, where annotators are asked to label each image $x \in \mathcal{U}$ with its accurate label $y \in \mathcal{C} := \{1,2,...,C\}$.
For each round $r$, with a maximum budget of $B$, we begin by constructing a candidate set of $B$ images from $\mathcal{U}_r$ utilizing class-guided clustering.
Each candidate is then evaluated with a class-wise threshold to determine whether to acquire ground-truth labels from annotators or use pseudo labels to conserve budget.
Finally, learnable prompts $t_r$ are trained by prompt learning on the dataset $\mathcal{D}_r$ accumulated up to round $r$.

In the followings, we introduce a class-guided clustering, which leverages the pre-trained image and text encoders of CLIP (Section~\ref{sec:multimodal-clustering}), and a cluster-balanced acquisition function to select candidate images (Section~\ref{sec:acquisition}).
After that, we introduce a budget-saving selective querying (Section~\ref{sec:budget-saving}). The overall algorithm is described in Algorithm~\ref{alg1}.

\subsection{Class-Guided Clustering}
\label{sec:multimodal-clustering}
Conventional active learning often relies on random sampling to construct the initial dataset, which can lead to the cold-start problem.
In contrast, we fully utilize CLIP's pre-trained image and text encoders for class-guided clustering, which can be applied at each round of the active learning process.
To construct class-guided features of an image $x$, we first define its image and text features, $I(x)$ and ${\tilde T}_{\mathcal{C}}(x)$, respectively.
For the image features, we simply use the image encoder of CLIP as follows:
\begin{align}
I(x) := \theta_{\text{img}}(x) \;.
\label{eq:image-features}
\end{align}
To derive the text features of $x$, we compute a weighted combination of text features with soft labels as weights:
\begin{align}
{\tilde T}_{\mathcal{C}}(x;t_{r-1}) := \sum_{c \in \mathcal{C}} p_\theta(y = c \mid x, t_{r-1}) \theta_{\text{txt}}(t_{r-1,c}) \;,
\label{eq:weighted-text}
\end{align}
where $t_{r-1}$ represents the learned vectors from the previous round, and $t_{r-1,c}$ concatenates $t_{r-1}$ with class $c$ as described in~\eqref{eq:learnable-vectors}.
For the initial round, we set $t_{r-1}$ as “a photo of a ”.
Finally, we concatenate the image and text features to obtain the class-guided features $\mathcal{F}_{\mathcal{C}}(x)$ for image $x$ as:
\begin{align}
\mathcal{F}_{\mathcal{C}}(x;t_{r-1}) := [ I(x), {\tilde T}_{\mathcal{C}}(x;t_{r-1}) ] \;.
\label{eq:multimodal-embeddings}
\end{align}
After that, we apply the $K$-means clustering algorithm~\citep{macqueen1967some} on the set of class-guided features across all images to partition them into $K$ clusters.


\vspace{1mm}
\noindent\textbf{GradFAM on class-guided features.}
\label{sec:heatmap}
We modify GradCAM~\citep{selvaraju2017gradcam} into GradFAM (Gradient-weighted Feature Activation Mapping) to visualize the influence of class-guided features on an image.
GradCAM highlights the degree to which individual pixels contribute to a specific class by analyzing the gradients of the class score with respect to the feature maps.
To tailor GradCAM for VLMs, we introduce the concept of \textit{target features} denoted by $\mathcal{F}_{\text{target}}$, allowing GradCAM to visualize the influence of target features on an image $x$ based on the cosine similarity score defined as $\text{cos} (I(x), \mathcal{F}_{\text{target}})$.
Our analysis technique, called GradFAM, visualizes the influence of target features rather than target class, offering the advantage of enabling label-free analysis.
Figure~\ref{subfig:heatmap-b} represents the case where $\mathcal{F}_{\text{target}} = I(x)$.
Consistent with CLIP's approach of embedding images and texts into a shared space, we observe that the influence of the target image features primarily concentrates on the overall objects within the image.
On the other hand, when the target features are guided by the set of classes, \ie $\mathcal{F}_{\text{target}} = \mathcal{F}_{\mathcal{C}}(x;t)$ where $t$ is “a photo of a ", Figures~\ref{subfig:heatmap-c} to~\ref{subfig:heatmap-f} distinctly highlight the specific objects corresponding to the guiding class sets.
To compute cosine similarity, we concatenate two copies of the same image features to match the dimensionality of the other features.
We note that our class-guided features incorporates class-relevant information through the text encoder, potentially resulting in clustering that aligns more closely with the target classifier, as described in Figure~\ref{fig:tsne-birds}.

\subsection{Cluster-Balanced Acquisition Function}
\label{sec:acquisition}
For ease of explanation, we first outline the data selection process in the initial round of active learning.
For a cluster-balanced acquisition, we set the number of clusters $K$ equal to the budget $B$, allowing for the selection of one image per cluster.
To choose the most representative sample from each cluster $C_i$, we first calculate its centroid $c_i$ as follows:
\begin{align}
c_i := \frac{1}{|C_i|} \sum_{x \in C_i} \mathcal{F}_{\mathcal{C}}(x;t) \;.
\end{align}
The closest image $x^*_i$ to the centroid $c_i$ is then selected as:
\begin{align}
x^*_i := \arg \min_{x \in C_i} ||\mathcal{F}_{\mathcal{C}}(x;t) - c_i||_2 \;.
\label{eq:closest}
\end{align}
We can construct the set of candidate images for querying:
\begin{align}
\mathcal{Q} := \{x^*_1, x^*_2, \dots, x^*_B \} \;.
\label{eq:candidates}
\end{align}
In the initial round, we consume the entire budget $B$ to request annotations for all candidate images in $\mathcal{Q}$.
Here, $t$ is replaced by $t_{r-1}$, and $\mathcal{Q}$ by $\mathcal{Q}_r$ for a subsequent round $r$.


\vspace{1mm}
\noindent\textbf{Subsequent rounds with increasing $K$.}
To enhance the diversity in the selected data, we introduce a progressively increasing $K$ based on round $r$, \ie $K = B \times r$.
This linear increase in $K$ ensures that at least $B$ clusters remain unlabeled in each round, allowing for the selection of clusters not included in previous rounds.
However, due to the inaccuracy of clustering in earlier rounds, samples previously assigned to different clusters may now be classified into the same cluster.
To address these cases, we prioritize larger clusters by sorting all unlabeled clusters by size and selecting the top-$K$ clusters.
Ablation studies on $K$ are in the Appendix~\ref{app:kmeans}.

\subsection{Selective Querying}
\label{sec:budget-saving}
In each round $r$, we can allocate the entire budget $B$ to acquire labels for all candidate images in $\mathcal{Q}_r$.
However, CLIP has demonstrated decent zero-shot performance in downstream classification tasks~\citep{radford2021learning} and performs even better with a few labeled samples~\citep{zhou2022learning}.
Therefore, for a candidate image where CLIP is already sufficiently confident in its label, we skip manual labeling and apply a pseudo label to conserve budget.

Since CLIP's knowledge is imbalanced across classes in classification tasks~\citep{bang2024active}, we propose a selective querying with class-wise thresholds.
To this end, we leverage the confidence scores of images from the previous training dataset $\mathcal{D}_{r-1}$.
In round $r$, the threshold $\epsilon_{r,c}$ for class $c \in \mathcal{C}$ is computed as follows:
\begin{align}
\epsilon_{r,c} := \frac{1}{|\mathcal{D}_{r-1,c}|} \sum_{(x, y=c) \in \mathcal{D}_{r-1,c}} p_\theta(y \mid x; t_{r-1}, \mathcal{C})\; ,
\label{eq:thresholds}
\end{align}
where $\mathcal{D}_{r-1,c}$ represents the subset of the training dataset labeled as $c$.
Note that thresholds for round $r$ depend on prior information, including $\mathcal{D}_{r-1}$ and $t_{r-1}$.
Therefore, a selective querying is impossible in the initial round.
We finally apply pseudo labels to candidates if their confidence exceeds the corresponding threshold.
As a result, the dataset $\mathcal{D}_r$ at round $r$ can be constructed as follows:
\begin{align}
\mathcal{D}_r := 
& \{(x, y) \mid x \in \mathcal{Q}_r, \; p_\theta(\tilde y \mid x ; t_{r-1}, \mathcal{C}) < \epsilon_{r,\tilde y} \} \cup \mathcal{D}_{r-1} \nonumber \\
\cup \; 
& \{(x, \tilde y) \mid x \in \mathcal{Q}_r, \; p_\theta(\tilde y \mid x; t_{r-1}, \mathcal{C}) \geq \epsilon_{r,\tilde y} \} \; ,
\label{eq:dataset}
\end{align}
where the pseudo label $\tilde y = y_\theta (x ; t, \mathcal{C})$ is defined as in~\eqref{eq:pseudo-label}.
Here, we note that $| \mathcal{D}_r | = B \times r$, yet the budget required is actually lower thanks to our selective querying.

To avoid bias in prompts trained during the previous round, we reinitialize $t_r$ randomly at each round and train prompts by minimizing the following CE loss:
\begin{align}
\hat{\mathbb{E}}_{(x, y) \sim {\mathcal{D}_r}} \big[ \text{CE} \big( y, p_\theta (y; x, t_r, \mathcal{C}) \big) \big] \;.
\label{eq:final-loss}
\end{align}

\noindent\textbf{Revisiting a unified prompt for prompt learning.}
Recent studies in prompt learning for VLMs have introduced sophisticated prompt designs, including image-wise~\citep{zhou2022conditional,yao2024tcp} and class-wise prompts~\citep{zhou2022learning, bang2024active,yao2024tcp}.
However, these prompts are prone to overfitting, particularly in active learning scenarios where only a limited number of samples are available.
To mitigate this issue, we introduce a new similarity measure that incorporates both a unified prompt $t_u$ and class-wise prompts $t_{\mathcal{C}} = \{t_{c} \mid c \in \mathcal{C} \}$.
Specifically, we define the cosine similarity between image $x$ and class $c$ as:
\begin{align}
\frac{
\text{cos} (\theta_{\text{img}}(x) , \theta_{\text{txt}}(t_{u,c})) + \text{cos} (\theta_{\text{img}}(x) , \theta_{\text{txt}}(t_{c}))}{2} \;,
\end{align}
where $t_{u,c}$ concatenates $t_u$ with class $c$.
We demonstrate that incorporating a unified prompt is beneficial in active learning scenarios with limited resources and enhances the effectiveness of the proposed selective querying.

\section{Experiments}
\label{sec:experiments}
\subsection{Experimental Setup}

\noindent\textbf{Datasets and implementation details.}
Following a previous study~\citep{bang2024active}, we use seven publicly available image classification datasets: OxfordPets (pet species)~\citep{parkhi2012cats}, FGVCAircraft (aircraft types)~\citep{maji2013fine}, Caltech101 (general object categories)~\citep{fei2004learning}, Flowers102 (flower species)~\citep{nilsback2008automated}, DTD (texture patterns)~\citep{cimpoi2014describing}, StanfordCars (car models)~\citep{krause20133d}, and EuroSAT (satellite land cover types)~\citep{helber2019eurosat}.
In our experiments, we employ CLIP ViT-B/32~\citep{dosovitskiy2021an,radford2021learning} as our VLM. 
At each round $r$, we reinitialize the learnable prompts $t_r$, consisting of 16 vectors, using a Gaussian distribution with a mean of 0 and a standard deviation of 0.02.
Following the training details in CoOp~\citep{zhou2022learning}, we train these prompts for 200 epochs per round using the SGD optimizer, initialized with a learning rate of 0.002 and decaying according to a cosine annealing.

\subsection{Active Learning Scenario}
\label{sec:active-learning-scenario}
\noindent\textbf{Baselines.}
Our cluster-balanced acquisition with selective querying (\sftype{CB}+\sftype{SQ})
is compared with the state-of-the-art (SOTA) active prompt learning method for VLMs, known as pseudo-class balance (\sftype{PCB})~\citep{bang2024active}, which operates based on~\sftype{BADGE}~\citep{ashdeep}.
In addition, we compare with conventional acquisitions commonly used in active learning for classification tasks, including~\sftype{Random},~\sftype{Entropy}~\cite{holub2008entropy} and~\sftype{CoreSet}~\citep{sener2018active}.
To ensure a fair reproduction of all baseline results, we follow the experimental setting of~\sftype{PCB}, leveraging class descriptions from LLMs~\citep{menonvisual} to train class-wise text prompts following CoOp~\citep{zhou2022learning}.

\vspace{1mm}
\noindent\textbf{Evaluation protocol.}
For a fair comparison, we follow the active learning scenario established in~\sftype{PCB}~\citep{bang2024active}.
Specifically, experiments are conducted over 8 rounds, with the maximum budget per round set to the number of classes, \ie $B = |\mathcal{C}|$.
Since the total budget varies across datasets, we set it to be fully spent at 100\% by the 8th round, with a maximum budget of 12.5\% per round.
Here, a budget of one indicates that an oracle assigns the ground-truth label for a single image.
Thanks to our selective querying in~\sftype{CB}+\sftype{SQ}, we generally consume less budget per round than other baselines from the second round onward.
We report the average accuracy over three trials, with shaded regions in the graphs representing the standard deviation.

\input{Figures_tex_main/fig_main_line_plot_iccv}

\vspace{1mm}
\noindent\textbf{Effect of proposed method.}
Figure~\ref{fig:main-graph} demonstrates the effectiveness of our proposed acquisition~\sftype{CB}+\sftype{SQ} across seven datasets.
While other baselines rely on~\sftype{Random} acquisition to mitigate the cold-start problem, \ie initial performance degradation compared to~\sftype{Random}, our~\sftype{CB}+\sftype{SQ} leverages the pre-trained image and text encoders of CLIP to enable a warm-start, leading to consistently strong early performance.
As shown in Figure~\ref{subfig:main-graph-a}, our method shows a 19.5\%p performance gain over the baselines at the first acquisition round.
Notably, with only $|\mathcal{C}|$ queried samples, our~\sftype{CB}+\sftype{SQ} already outperforms other baselines trained with $3|\mathcal{C}|$ samples, highlighting its sample efficiency.
In addition, our selective querying allows~\sftype{CB}+\sftype{SQ} to outperform other baselines while reducing the labeling budget by 17.6\%.
These gains are consistent across both fine-grained and general datasets, and we also observe the same phenomenon reported in~\cite{bang2024active}, where conventional acquisition methods such as~\sftype{Entropy} and~\sftype{CoreSet} underperform even~\sftype{Random} sampling.

\subsection{Further Analyses}
\label{sec:further-analyses}

\noindent\textbf{Extensions to SOTA model-centric prompt learning methods.}
Our cluster-balanced (\sftype{CB}) acquisition effectively selects informative images by utilizing a pre-trained CLIP model.
These selected images can be directly applied to enhance the performance of SOTA model-centric prompt learning methods, including MaPle~\citep{khattak2023maple}, PromptSRC~\citep{khattak2023self}, and ProMetaR~\citep{park2024prompt}.
Table~\ref{tab:main-table} shows that our~\sftype{CB}-based datasets slightly outperforms the conventional~\sftype{1-shot} datasets, which contain one labeled image per class.
More specifically, following the few-shot setting of CoOp~\citep{zhou2022learning}, we assume access to ground-truth labels to construct a perfectly class-balanced~\sftype{1-shot} dataset by randomly selecting one sample from each class.
However, the~\sftype{1-shot} datasets differ from active learning, which start with unlabeled images.
For a fair comparison with them, we construct~\sftype{CB}*-based datasets, where the weighted text features in~\eqref{eq:weighted-text} are replaced with the text features of the ground-truth label.
In Table~\ref{tab:main-table}, our~\sftype{CB}*-based datasets outperform other baselines, emphasizing the importance of data-centric approaches for efficient adaptation in VLMs.

\begin{table*}[!t]
    \setlength\tabcolsep{4pt}
    \centering
    \footnotesize
    \caption{{\em Synergy of the proposed acquisition with existing prompt learning methods.} All baseline methods are trained with~\sftype{1-shot} datasets. Our~\sftype{CB}*-based curated datasets enhance the performance of previous model-centric prompt learning methods.}
    \label{tab:main-table}
    
    \begin{tabular}{lcccccccc}
    \toprule
    \textbf{Methods} & \textbf{Flowers102} & \textbf{DTD} & \textbf{OxfordPets} & \textbf{StanfordCars} & \textbf{Caltech101} & \textbf{Aircraft} & \textbf{EuroSAT} & \textbf{Average (\%)} \\
    
    \midrule
    MaPle & 75.23 & 48.77 & 85.27 & 57.30 & 91.57 & 18.33 & 61.67 & 62.59 \\
    \ + CB & 78.80 & 46.47 & 87.23 & 55.70 & 90.17 & 16.00 & \textbf{73.50} & \underline{63.98} \\
    \rowcolor{Gray}
    \ + CB* & \textbf{80.70} & \textbf{51.07} & \textbf{88.83} & \textbf{60.47} & \textbf{93.07} & \textbf{20.85} & 71.07 & \textbf{66.58} \\
    \midrule
    PromptSRC & 77.60 & 51.53 & 89.60 & 63.67 & 93.10 & 18.67 & 66.93 & 65.87 \\
    \ + CB & 79.73 & 50.87 & 89.80 & 61.30 & 92.40 & \textbf{22.03} & 70.20 & \underline{66.62} \\
    \rowcolor{Gray}
    \ + CB* & \textbf{80.43} & \textbf{56.17} & \textbf{90.13} & \textbf{63.97} & \textbf{93.97} & 21.27 & \textbf{70.33} & \textbf{68.04} \\
    \midrule
    ProMetaR & 78.87 & 52.07 & 87.63 & 60.60 & 92.53 & 19.50 & 68.13 & 65.62 \\
    \ + CB & 80.43 & 49.10 & 89.53 & 59.37 & 91.80 & 20.37 & 71.80 & \underline{66.06} \\
    \rowcolor{Gray}
    \ + CB* & \textbf{83.37} & \textbf{55.20} & \textbf{89.53} & \textbf{63.83} & \textbf{93.90} & \textbf{22.10} & \textbf{72.87} & \textbf{68.69} \\
    
    \bottomrule
    \vspace{1.5mm}
    \end{tabular}
\end{table*}

\begin{figure*}[t!]
    \centering
    \begin{subfigure}[t]{0.245\textwidth}
        \centering
        \begin{tikzpicture}
            \node[anchor=north west, inner sep=0] (image) at (0, 0) {\fbox{\includegraphics[width=0.8\linewidth, height=0.8\linewidth]{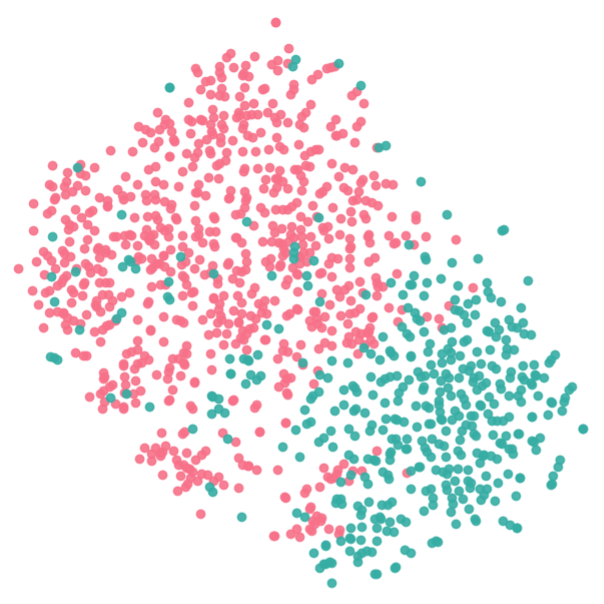}}};
            \node[anchor=north west] at (image.north west) {\tiny ARI: 45.27\%};
        \end{tikzpicture}
        \caption{Image features}
        \label{subfig:tsne-birds-a}
    \end{subfigure} %
    \hfill
    \begin{subfigure}[t]{0.245\textwidth}
        \centering
        \begin{tikzpicture}
            \node[anchor=north west, inner sep=0] (image) at (0, 0) {\fbox{\includegraphics[width=0.8\linewidth, height=0.8\linewidth]{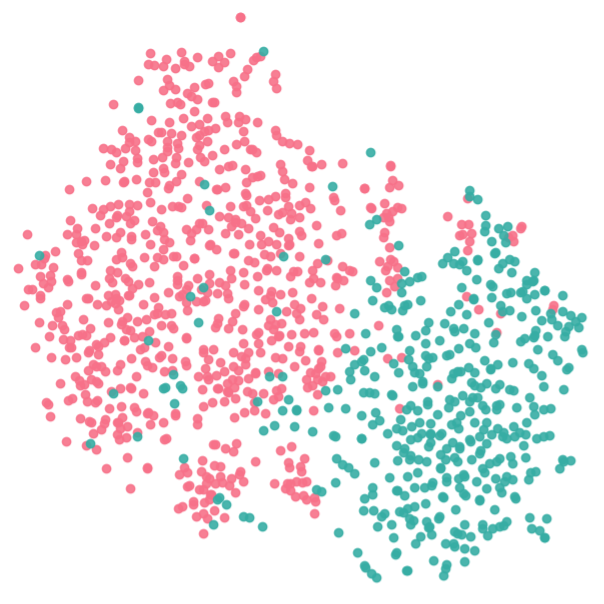}}};
            \node[anchor=north west] at (image.north west) {\tiny ARI: 56.57\%};
        \end{tikzpicture}
        \caption{Class-guided features}
        \label{subfig:tsne-birds-b}
    \end{subfigure} %
    \hfill
    \begin{subfigure}[t]{0.245\textwidth}
        \centering
        \begin{tikzpicture}
            \node[anchor=north west, inner sep=0] (image) at (0, 0) {\fbox{\includegraphics[width=0.8\linewidth, height=0.8\linewidth]{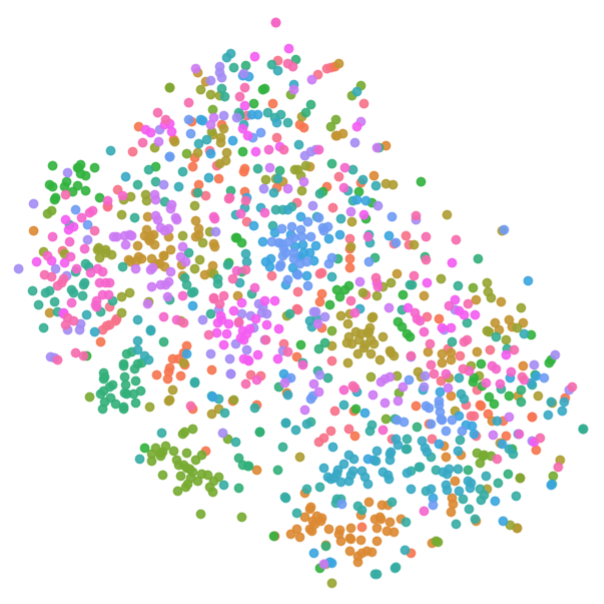}}};
            \node[anchor=north west] at (image.north west) {\tiny ARI: 12.74\%};
        \end{tikzpicture}
        \caption{Image features}
        \label{subfig:tsne-birds-c}
    \end{subfigure} %
    \hfill
    \begin{subfigure}[t]{0.245\textwidth}
        \centering
        \begin{tikzpicture}
            \node[anchor=north west, inner sep=0] (image) at (0, 0) {\fbox{\includegraphics[width=0.8\linewidth, height=0.8\linewidth]{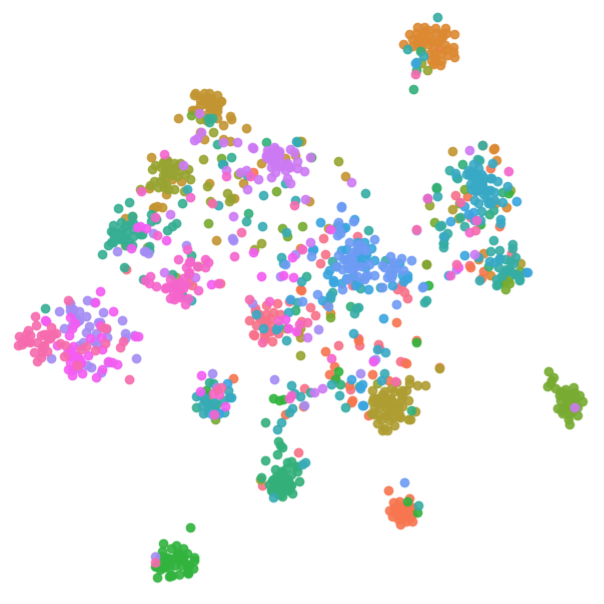}}};
            \node[anchor=north west] at (image.north west) {\tiny ARI: 37.64\%};
        \end{tikzpicture}
        \caption{Class-guided features}
        \label{subfig:tsne-birds-d}
    \end{subfigure}
    \caption{{\em T-SNE for class-guided clustering.} (a, c) Clustering based solely on image features results in clusters that are poorly separated. (a) and (c) correspond to the $|\mathcal{C}|=2$ and $|\mathcal{C}|=20$ settings, respectively. (b, d) In contrast, our class-guided clustering leads to more distinct clusters that align with the size of the guiding class set $\mathcal{C}$.}
    \label{fig:tsne-birds}
\end{figure*}

\noindent\textbf{Visualization for class-guided clustering.}
In Section~\ref{sec:heatmap}, we analyze the proposed class-guided features using GradFAM, as shown in Figure~\ref{fig:heatmap}.
Here, we further visualize the impact of class-guided features on clustering with T-SNE~\citep{van2008visualizing} on the Waterbirds dataset~\citep{sagawadistributionally}.
Specifically, the Waterbirds dataset consists of 200 distinct bird species, where each image is labeled by habitat (water, land), background (water, land), and specific species.
This allows us to categorize the dataset into 2, 4, or 200 groups.
For improved visual clarity, we analyze a subsample of 20 classes.
Figures~\ref{subfig:tsne-birds-a} and~\ref{subfig:tsne-birds-c}  shows that the limitation of relying solely on image features in capturing semantic information.
In contrast, our class-guided features, which incorporate the set of classes $\mathcal{C}$, effectively reflect relevant class information in the clustering results, as illustrated in Figures~\ref{subfig:tsne-birds-b} and~\ref{subfig:tsne-birds-d}.
This improvement is also reflected in the Adjusted Rand Index (ARI), where the score increases from 12.74\% to 37.64\% when moving from image-only features to class-guided features.
More details are provided in Appendix~\ref{sec:waterbirds-details}.


\input{Figures_tex_main/fig_selective_querying}

\begin{figure}[!t]
    \centering
    
    \pgfplotstableread[col sep=comma]{Data/StanfordCars_4_confidence_counts_50.csv}\stepData
    \pgfplotstableread[col sep=comma]{Data/StanfordCars_4_confidence_mean.csv}\meanData
    
    \pgfplotstablegetelem{0}{csctrue_labeled}\of\meanData
    \pgfmathsetmacro{\meanCscTrueLabeled}{\pgfplotsretval}
    
    \pgfplotstablegetelem{0}{csctrue_unlabeled}\of\meanData
    \pgfmathsetmacro{\meanCscTrueUnlabeled}{\pgfplotsretval}
    
    \pgfplotstablegetelem{0}{cscfalse_labeled}\of\meanData
    \pgfmathsetmacro{\meanCscFalseLabeled}{\pgfplotsretval}
    
    \pgfplotstablegetelem{0}{cscfalse_unlabeled}\of\meanData
    \pgfmathsetmacro{\meanCscFalseUnlabeled}{\pgfplotsretval}
    


    \begin{subfigure}[t]{0.45\linewidth}
        \centering
        \begin{tikzpicture}
        \begin{axis}[
            width=1\linewidth,
            height=0.7\linewidth,
            label style={font=\scriptsize},
            tick label style={font=\scriptsize},
            xmin=0, xmax=1.0,  
            ymin=0, ymax=40,
            ylabel={Count},
            xlabel={$p(y \mid x; t_{\mathcal{C}}, \mathcal{C})$},
            xlabel style={yshift=0.15cm},
            ylabel style={yshift=-0.6cm},
            mark size=1.4pt,
            legend style={
                nodes={scale=0.6}, 
                at={(0.29, 0.97)}, 
                /tikz/every even column/.append style={column sep=1mm},
                legend columns=1,
                legend image post style={xscale=0.5, yscale=0.5, solid}
            },
        ]
            \addplot[
                const plot,
                draw=cRed,
                line width=1.2pt
            ] table [x=x, y=csctrue_unlabeled] {\stepData};
            \addlegendentry{Untrained}
            
            \addplot[
                const plot,
                draw=cBlue,
                line width=1.2pt
            ] table [x=x, y=csctrue_labeled] {\stepData};
            \addlegendentry{Trained}
    
            \addplot[
                const plot,
                draw=cRed,
                line width=1.2pt
            ] table [x=x, y=csctrue_unlabeled] {\stepData};
            
            \draw[densely dotted, color=cBlue, thick]
                (axis cs:\meanCscTrueLabeled, 0)
                -- (axis cs:\meanCscTrueLabeled, \pgfkeysvalueof{/pgfplots/ymax} + 1);
            
            \draw[densely dotted, color=cRed, thick]
                (axis cs:\meanCscTrueUnlabeled, 0)
                -- (axis cs:\meanCscTrueUnlabeled, \pgfkeysvalueof{/pgfplots/ymax} + 1);
        \end{axis}
        \end{tikzpicture}
        \caption{Class-wise prompts ($t_{\mathcal{C}}$)}
    \end{subfigure}
    \begin{subfigure}[t]{0.45\linewidth}
        \centering
        \begin{tikzpicture}
        \begin{axis}[
            width=1\linewidth,
            height=0.7\linewidth,
            label style={font=\scriptsize},
            tick label style={font=\scriptsize},
            xmin=0, xmax=1.0,
            ymin=0, ymax=40,
            xlabel={$p(y \mid x; t_{u}, \mathcal{C})$},
            ylabel={Count},
            xlabel style={yshift=0.15cm},
            ylabel style={yshift=-0.6cm},
            legend style={
                nodes={scale=0.6}, 
                at={(0.29, 0.97)}, 
                /tikz/every even column/.append style={column sep=1mm},
                legend columns=1,
                legend image post style={xscale=0.5, yscale=0.5, solid}
            },
        ]
            \addplot[
                const plot,
                draw=cRed,
                line width=1.2pt
            ] table [x=x, y=cscfalse_unlabeled] {\stepData};
            \addlegendentry{Untrained}
            
            \addplot[
                const plot,
                draw=cBlue,
                line width=1.2pt
            ] table [x=x, y=cscfalse_labeled] {\stepData};
            \addlegendentry{Trained}
            
            \addplot[
                const plot,
                draw=cRed,
                line width=1.2pt
            ] table [x=x, y=cscfalse_unlabeled] {\stepData};
    
            \draw[densely dotted, color=cBlue, thick]
                (axis cs:\meanCscFalseLabeled, 0)
                -- (axis cs:\meanCscFalseLabeled, \pgfkeysvalueof{/pgfplots/ymax});
    
            \draw[densely dotted, color=cRed, thick]
                (axis cs:\meanCscFalseUnlabeled, 0)
                -- (axis cs:\meanCscFalseUnlabeled, \pgfkeysvalueof{/pgfplots/ymax});
        \end{axis}
        \end{tikzpicture}
        \caption{Unified prompt ($t_u$)}
    \end{subfigure}
    
    
    \caption{{\em Confidence distributions.} Due to overfitting in (a) the class-wise prompts, the overall confidence scores and the mean confidence (dotted line) of trained samples are significantly higher than those obtained with (b) the unified prompt.}
    \label{fig:confidence}
\end{figure}

\vspace{1mm}
\noindent\textbf{Synergy of selective querying with our method.}
Our selective querying (\sftype{SQ}) synergizes strongly with our method for two key reasons.
First, our cluster-balanced acquisition follows a diversity-driven strategy, which may occasionally select well-understood samples.
In such cases, \sftype{SQ} leverages pseudo labels to conserve budget.
As shown in Figure~\ref{fig:effect-sq-a}, \sftype{SQ} is particularly effective in diversity-based methods such as~\sftype{CoreSet} and~\sftype{CB}, while uncertainty-based approaches like~\sftype{Entropy} and~\sftype{PCB} show limited improvement in Figure~\ref{fig:effect-sq-b}.
Second, we incorporate a unified prompt, as described in Section~\ref{sec:budget-saving}, further enhancing the effectiveness of \sftype{SQ}.
Figure~\ref{fig:confidence} shows that the unified prompt results in confidence scores for trained samples with less overfitting compared to class-wise prompts, leading to more reliable class-wise thresholds in~\eqref{eq:thresholds}.
This observation aligns with the results in Figure~\ref{fig:prompt-types}, showing that a unified prompt is more effective in reducing the labeling budget than class-wise prompts.

\vspace{1mm}
\noindent\textbf{Confidence distributions by prompt type.}
\label{app:confi}
Figure \ref{fig:confidence} compares the confidence distributions of trained and untrained samples. We first train either class-wise prompts $t_{\mathcal{C}}$ or a unified prompt $t_u$ on the StanfordCars dataset $\mathcal{D}$ \citep{krause20133d}.
To emulate an active-learning setting with a limited budget, we form a 4-shot training subset $\mathcal{D}_t \subset \mathcal{D}$.
For each class $c$, we then compute the average confidence score $s_{t,c}$ as follows:
\begin{align}
s_{t,c} := \frac{1}{|\mathcal{D}_{t,c}|} \sum_{(x,y) \in \mathcal{D}_{t,c}} p(y \mid x; t, \mathcal{C}) \;,
\label{eq:c}
\end{align}
where $\mathcal{D}_{t,c}$ is the portion of $\mathcal{D}_t$ that belongs to class $c$, and $t$ is either $t_{\mathcal{C}}$ or $t_u$.
We then plot 50-bin histograms of these scores, using blue for the training subset $\mathcal{D}_t$ and red for the remaining data $\mathcal{D} \setminus \mathcal{D}_t$.
The resulting plots show that class-wise prompts concentrate confidence on seen samples, which may indicate overfitting, whereas the unified prompt yields a more balanced confidence distribution across unseen data.



\vspace{1mm}
\noindent\textbf{Effect of various feature spaces.}
In Table~\ref{tab:diff_feat}, we examine the effect of various feature spaces on our cluster-balanced acquisition, as an alternative to the proposed class-guided features that leverage both~\sftype{Image} and~\sftype{Text} features.
Table~\ref{tab:diff_feat} demonstrates that our class-guided features yield a 3.2\%p and 7.7\%p improvement over using only~\sftype{Image} and only~\sftype{Text} features, respectively.
In addition, we experiment with a conventional few-shot labeled dataset that relies solely on~\sftype{Labels} for a perfectly class-balanced dataset, which can be impractical in an active learning scenario.
For a fair comparison, we redefine the weighted text features in~\eqref{eq:weighted-text} as the text features of ground-truth labels, resulting in an average performance increase of 10.2\%p.

\vspace{1mm}
\noindent\textbf{Analyses of constructed datasets.}
A primary goal of active learning is to construct high-quality datasets with minimal labeling budgets.
Figure~\ref{fig:analyses-datasets} demonstrates that our method builds high-quality datasets with a significantly smaller budget compared to previous methods.
Specifically, while previous methods consume the entire budget each round to generate a 100\% clean dataset, our selective querying conserves budget by using pseudo-labels based on class-wise thresholds, yet still achieves a comparable level of dataset quality.
This results in performance on par with previous methods, as shown in Figure~\ref{fig:main-graph}.
Additionally, we note that as rounds progress, the budget-saving advantage of our method becomes more pronounced.

\begin{table}[t!]
\caption{{\em Ablation studies on different feature spaces.} In the initial round, our class-guided features (third row), leveraging both image and text features, demonstrate effectiveness across 7 datasets.}
\label{tab:diff_feat}
\begin{center}
\setlength\tabcolsep{6pt}
\centering
\footnotesize
\begin{tabular}{cccc}
\toprule
\textbf{Image Features} & \textbf{Text Features} & \textbf{Ground-truth} & \textbf{Average Acc. (\%)} \\ \midrule
\cmark & \xmark & \xmark & 52.72$_{\pm 0.37}$ \\
\xmark & \cmark$_{\text{soft }}$ & \xmark & 48.24$_{\pm 0.21}$ \\ 
\rowcolor{Gray}
\cmark & \cmark$_{\text{soft }}$ & \xmark & $\textbf{55.92}_{\pm 0.38}$ \\ \midrule
\xmark & \xmark & \cmark & 48.10$_{\pm 0.41}$ \\ 
\rowcolor{Gray}
\cmark & \cmark$_{\text{label}}$ & \cmark & $\textbf{58.27}_{\pm 0.43}$ \\ 
\bottomrule
\end{tabular}
\end{center}
\end{table}

\begin{figure}[t]
  \noindent
  \begin{minipage}[t]{0.6\linewidth}
    \centering
    \begin{subfigure}[t]{0.45\linewidth}
      \centering
      \begin{tikzpicture}
        \begin{axis}[
          scale only axis,
          enlargelimits=false,
          clip mode=individual,
          label style={font=\scriptsize},
          tick label style={font=\scriptsize},
          width=0.85\linewidth,
          height=0.89\linewidth,
          xlabel=Rounds,
          ylabel=Ratio (\%),
          xmin=0.5, xmax=8.5,
          ymin=80, ymax=103,
          xtick={1, 2, 3, 4, 5, 6, 7, 8},
          ytick={75, 80, 85, 90, 95, 100},
          xlabel style={yshift=0.15cm},
          ylabel style={yshift=-0.6cm},
          mark size=1.4pt,
          legend style={
            nodes={scale=0.6},
            at={(0.42,0.2)},
            /tikz/every even column/.append style={column sep=1mm},
            legend columns=1,
            legend image post style={
              mark size=2.4pt, xscale=0.5, yscale=0.5, solid}
          },
        ]
          \addplot[cPink, thick, mark=*, mark options={solid}]
            table[col sep=comma, x=x, y=Empty]
            {Data/Exp_Pseudo-label_Accuracy_iccv.csv};

          \addplot[gray, thick, mark=square*, mark options={solid}]
            table[col sep=comma, x=x, y=Empty]
            {Data/Exp_Pseudo-label_Accuracy_iccv.csv};

          \addplot[gray, thick, mark=square*, mark options={solid}]
            table[col sep=comma, x=x, y=Baselines]
            {Data/Exp_Pseudo-label_Accuracy_iccv.csv};

          \addplot[cPink, thick, mark=*, mark options={solid}]
            table[col sep=comma, x=x, y=Consumed-budget-ratio]
            {Data/Exp_Pseudo-label_Accuracy_iccv.csv};

          \legend{CB+SQ, Baselines}
        \end{axis}
      \end{tikzpicture}
      \caption{Required budget}
      \label{pseudo-a}
    \end{subfigure}
    \hspace{3mm}
    \begin{subfigure}[t]{0.45\linewidth}
      \centering
      \begin{tikzpicture}
        \begin{axis}[
          scale only axis,
          enlargelimits=false,
          clip mode=individual,
          label style={font=\scriptsize},
          tick label style={font=\scriptsize},
          width=0.85\linewidth,
          height=0.89\linewidth,
          xlabel=Rounds,
          ylabel=Correct label ratio (\%),
          xmin=0.5, xmax=8.5,
          ymin=80, ymax=103,
          xtick={1, 2, 3, 4, 5, 6, 7, 8},
          ytick={75, 80, 85, 90, 95, 100},
          xlabel style={yshift=0.15cm},
          ylabel style={yshift=-0.6cm},
          mark size=1.4pt,
          legend style={
            nodes={scale=0.6},
            at={(0.42,0.2)},
            /tikz/every even column/.append style={column sep=1mm},
            legend columns=1,
            legend image post style={
              mark size=2.4pt, xscale=0.5, yscale=0.5, solid}
          },
        ]
          \addplot[cPink, thick, mark=*, mark options={solid}]
            table[col sep=comma, x=x, y=Dataset-quality]
            {Data/Exp_Pseudo-label_Accuracy_iccv.csv};

          \addplot[gray, semithick, mark=square*, mark options={solid}]
            table[col sep=comma, x=x, y=Baselines]
            {Data/Exp_Pseudo-label_Accuracy_iccv.csv};

          \legend{CB+SQ, Baselines}
        \end{axis}
      \end{tikzpicture}
      \caption{Dataset quality}
      \label{pseudo-b}
    \end{subfigure}
    \caption{{\em Analyses of constructed datasets.} (a) As rounds progress, we construct datasets with progressively smaller budgets. (b) The quality of the constructed dataset remains consistently high, regardless of the number of rounds.}
    \label{fig:analyses-datasets}
  \end{minipage}
  \hspace{4mm}
  \begin{minipage}[t]{0.33\linewidth}
    \begin{subfigure}[t]{\linewidth}
    \centering
    \begin{tikzpicture}
      \begin{axis}[
        scale only axis,
        enlargelimits=false,
        clip mode=individual,
        label style={font=\scriptsize},
        tick label style={font=\scriptsize},
        width=0.69\linewidth,
        height=0.723\linewidth,
        xlabel=Cumulative Budget (\%),
        ylabel=Accuracy (\%),
        xmin=6.25, xmax=106.25,
        ymin=23, ymax=65,
        xtick={0,25,50,75,100},
        ytick={15,25,35,45,55,65},
        xlabel style={yshift=0.15cm},
        ylabel style={yshift=-0.6cm},
        mark size=1.4pt,
        legend style={
          nodes={scale=0.6},
          at={(0.97,0.27)},
          /tikz/every even column/.append style={column sep=1mm},
          legend columns=1,
        }
      ]
        \addplot[cPink, thick, mark=*, mark options={solid}]
          table[col sep=comma, x=x, y=CBBS]
          {Data/rebuttal/rebuttal_iccv_b.csv};

        \addplot[cRed, thick, mark=triangle*, mark options={solid}]
          table[col sep=comma, x=x, y=Random]
          {Data/rebuttal/rebuttal_iccv_b.csv};

        \addplot[orange, thick, mark=diamond*, mark options={solid}]
          table[col sep=comma, x=x, y=Entropy]
          {Data/rebuttal/rebuttal_iccv_b.csv};

        \legend{CB+SQ, Random, Entropy}
      \end{axis}
    \end{tikzpicture}
    \caption{ImageNet}
    \label{pseudo-c}
    \end{subfigure}
    \caption{{\em ImageNet experiments.} Our approach is applicable to ImageNet and outperforms existing methods.}
    \label{reb:imagenet}
\end{minipage}
\end{figure}

\vspace{1mm}
\noindent\textbf{ImageNet experiments.}
ImageNet~\citep{deng2009imagenet} contains 1.28 M training images, making large-scale active learning computationally demanding.
\citet{emam2021active} reports that \sftype{CoreSet} and \sftype{BADGE} are infeasible at this scale, and that \sftype{PCB}, which builds on \sftype{BADGE}, suffers the same limitation. 
Our method overcomes this bottleneck by employing a lightweight $K$-means clustering step.
As shown in Figure~\ref{reb:imagenet}, the resulting \sftype{CB+SQ} strategy scales efficiently to ImageNet and achieves higher accuracy than all competing baselines.



\begin{table}[t!]
\begin{center}
\caption{{\em Effectiveness on unseen classes.} Our~\sftype{CB} shows strong performance on novel classes in the initial round across \khy{9 datasets.}}
\label{tab:base-to-novel}
\setlength\tabcolsep{8pt}
\centering
\footnotesize
\begin{tabular}{cccc}
\toprule
\textbf{Methods} & \textbf{Base} & \textbf{Novel} & \textbf{HM} \\ \midrule
Random & 59.02$_{\pm 1.06}$ & 58.94$_{\pm 1.48}$ & 58.98$_{\pm 1.14}$ \\ 
\rowcolor{Gray}
CB & \textbf{68.81}$_{\pm 1.07}$ & \textbf{63.68}$_{\pm 1.25}$ & \textbf{66.15}$_{\pm 0.67}$ \\
\bottomrule
\end{tabular}
\vspace{1mm}
\end{center}
\end{table}

\vspace{1mm}
\noindent\textbf{Base-to-novel generalization.}
Following previous work~\citep{zhou2022learning}, we divide each dataset's classes into base and novel groups. We then use acquisition functions to select a subset of unlabeled images from the base classes to train prompts, which are subsequently applied to the novel classes for evaluation.
In Table~\ref{tab:base-to-novel}, we compare our cluster-balanced (\sftype{CB}) acquisition with~\sftype{Random} reflecting the cold-start of other acquisitions in Figure~\ref{fig:main-graph}.
Table~\ref{tab:base-to-novel} demonstrates that our~\sftype{CB} acquisition outperforms~\sftype{Random} in both base and novel groups.
Here, HM denotes the harmonic mean of base and novel performance. 

\vspace{1mm}
\noindent\textbf{Various ablation studies.}
\label{sec:proposed-components}
Figure~\ref{subfig:ablation-a} examines the effect of different prompt types, including unified (\sftype{U}), class-wise (\sftype{C}), and both (\sftype{B}), revealing that our~\sftype{CB} consistently outperforms~\sftype{PCB} regardless of the prompt type.
In addition, we analyze the contribution of the proposed components: (i) selective querying (\sftype{SQ}) and (ii) the unified prompt (\sftype{C} to \sftype{B}), as illustrated in Figures~\ref{subfig:ablation-b} and~\ref{subfig:ablation-c}.
Although~\sftype{CoreSet} has demonstrated low performance in active learning for VLMs~\citep{bang2024active}, this is primarily due to its reliance solely on image features.
We observe that incorporating our class-guided features can enhance its performance to some extent as shown in Figure~\ref{subfig:ablation-c}, though it still fails to surpass~\sftype{PCB}.
We note that active learning for conventional neural networks with image-only encoders may differ from active learning for VLMs.

\input{Figures_tex_main/fig_coreset_iccv}

\section{Conclusion}
In this work, we propose an active prompt learning framework that leverages the prior knowledge of vision-language models (VLMs) for efficient data acquisition and prompt adaptation. Using the pre-trained image and text encoders of CLIP, we extract class-guided features that combine image embeddings with weighted text embeddings based on similarity scores, enabling K-means clustering to select diverse and representative samples. To further enhance budget efficiency, we introduce a selective querying strategy with adaptive class-wise thresholds, assigning pseudo-labels to high-confidence samples while requesting annotations for uncertain ones. Experiments across seven datasets and large-scale settings such as ImageNet show that our cluster-balanced acquisition with selective querying reduces annotation costs and achieves superior accuracy compared to baselines. Furthermore, our data-centric approach complements existing model-centric prompt learning methods, offering a general strategy for scalable VLM adaptation.

\noindent\textbf{Limitations and future work.}
Our framework relies on strong pre-trained vision-language models such as CLIP, which may limit its effectiveness when applied to weaker or domain-specific backbones. Moreover, it has so far been validated only on image classification tasks, leaving its applicability to other vision tasks. Future work will explore extending the approach to object detection, semantic segmentation, and real human-in-the-loop settings to assess its broader utility.

\vspace{1mm}
\noindent\textbf{Acknowledgements.}
This work was partly supported by the IITP grants and the NRF grants funded by Ministry of Science and ICT, Korea (No.RS-2019-II191906, Artificial Intelligence Graduate School Program (POSTECH); No.RS-2021-II212068, Artificial Intelligence Innovation Hub; No.RS-2023-00217286; No.RS-2021-II210739, Development of Distributed/Cooperative AI based 5G+ Network Data Analytics Functions and Control Technology; No.RS-2024-00457882, AI Research Hub Project).

\bibliography{main}

@String(ICCV= {Int. Conf. Comput. Vis.})

@String(ECCV= {Eur. Conf. Comput. Vis.})

@String(ICME = {Int. Conf. Multimedia and Expo})

@String(ICLR = {Int. Conf. Learn. Represent.})

@String(ICCV  = {ICCV})

@String(ECCV  = {ECCV})

@String(ICME  =	{ICME})

@String(ICLR  = {ICLR})

@inproceedings{menonvisual,
  title={Visual Classification via Description from Large Language Models},
  author={Menon, Sachit and Vondrick, Carl},
  booktitle={The Eleventh International Conference on Learning Representations},
  year={2022}
}

@inproceedings{yao2024tcp,
  title={Tcp: Textual-based class-aware prompt tuning for visual-language model},
  author={Yao, Hantao and Zhang, Rui and Xu, Changsheng},
  booktitle={Proceedings of the IEEE/CVF Conference on Computer Vision and Pattern Recognition},
  pages={23438--23448},
  year={2024}
}

@inproceedings{radford2021learning,
  title={Learning transferable visual models from natural language supervision},
  author={Radford, Alec and Kim, Jong Wook and Hallacy, Chris and Ramesh, Aditya and Goh, Gabriel and Agarwal, Sandhini and Sastry, Girish and Askell, Amanda and Mishkin, Pamela and Clark, Jack and others},
  booktitle={International conference on machine learning},
  pages={8748--8763},
  year={2021},
  organization={PMLR}
}

@article{zhou2022learning,
  title={Learning to prompt for vision-language models},
  author={Zhou, Kaiyang and Yang, Jingkang and Loy, Chen Change and Liu, Ziwei},
  journal={International Journal of Computer Vision},
  volume={130},
  number={9},
  pages={2337--2348},
  year={2022},
  publisher={Springer}
}

@inproceedings{kimactive,
  title={Active Label Correction for Semantic Segmentation with Foundation Models},
  author={Kim, Hoyoung and Hwang, Sehyun and Kwak, Suha and Ok, Jungseul},
  booktitle={Forty-first International Conference on Machine Learning},
  year={2024}
}

@article{bayer2024activellm,
  title={ActiveLLM: Large Language Model-based Active Learning for Textual Few-Shot Scenarios},
  author={Bayer, Markus and Reuter, Christian},
  journal={arXiv preprint arXiv:2405.10808},
  year={2024}
}

@inproceedings{macqueen1967some,
  title={Some methods for classification and analysis of multivariate observations},
  author={MacQueen, James and others},
  booktitle={Proceedings of the fifth Berkeley symposium on mathematical statistics and probability},
  year={1967},
  organization={Oakland, CA, USA}
}

@inproceedings{sagawadistributionally,
  title={Distributionally Robust Neural Networks},
  author={Sagawa, Shiori and Koh, Pang Wei and Hashimoto, Tatsunori B and Liang, Percy},
  booktitle={International Conference on Learning Representations},
  year={2019}
}

@inproceedings{bang2024active,
  title={Active Prompt Learning in Vision Language Models},
  author={Bang, Jihwan and Ahn, Sumyeong and Lee, Jae-Gil},
  booktitle={Proceedings of the IEEE/CVF Conference on Computer Vision and Pattern Recognition},
  pages={27004--27014},
  year={2024}
}

@inproceedings{liimage,
  title={Image Clustering with External Guidance},
  author={Li, Yunfan and Hu, Peng and Peng, Dezhong and Lv, Jiancheng and Fan, Jianping and Peng, Xi},
  booktitle={Forty-first International Conference on Machine Learning},
  year={2024}
}

@inproceedings{
ashdeep,
title={Deep Batch Active Learning by Diverse, Uncertain Gradient Lower Bounds},
author={Jordan T. Ash and Chicheng Zhang and Akshay Krishnamurthy and John Langford and Alekh Agarwal},
booktitle={International Conference on Learning Representations},
year={2020},
url={https://openreview.net/forum?id=ryghZJBKPS}
}

@inproceedings{sener2018active,
  title={Active Learning for Convolutional Neural Networks: A Core-Set Approach},
  author={Sener, Ozan and Savarese, Silvio},
  booktitle={International Conference on Learning Representations},
  year={2018}
}

@inproceedings{khattak2023maple,
  title={Maple: Multi-modal prompt learning},
  author={Khattak, Muhammad Uzair and Rasheed, Hanoona and Maaz, Muhammad and Khan, Salman and Khan, Fahad Shahbaz},
  booktitle={Proceedings of the IEEE/CVF Conference on Computer Vision and Pattern Recognition},
  pages={19113--19122},
  year={2023}
}

@inproceedings{khattak2023self,
  title={Self-regulating prompts: Foundational model adaptation without forgetting},
  author={Khattak, Muhammad Uzair and Wasim, Syed Talal and Naseer, Muzammal and Khan, Salman and Yang, Ming-Hsuan and Khan, Fahad Shahbaz},
  booktitle={Proceedings of the IEEE/CVF International Conference on Computer Vision},
  pages={15190--15200},
  year={2023}
}

@inproceedings{park2024prompt,
  title={Prompt Learning via Meta-Regularization},
  author={Park, Jinyoung and Ko, Juyeon and Kim, Hyunwoo J},
  booktitle={Proceedings of the IEEE/CVF Conference on Computer Vision and Pattern Recognition},
  pages={26940--26950},
  year={2024}
}

@inproceedings{holub2008entropy,
  title={Entropy-based active learning for object recognition},
  author={Holub, Alex and Perona, Pietro and Burl, Michael C},
  booktitle={2008 IEEE Computer Society Conference on Computer Vision and Pattern Recognition Workshops},
  pages={1--8},
  year={2008},
  organization={IEEE}
}

@inproceedings{cimpoi2014describing,
  title={Describing textures in the wild},
  author={Cimpoi, Mircea and Maji, Subhransu and Kokkinos, Iasonas and Mohamed, Sammy and Vedaldi, Andrea},
  booktitle={Proceedings of the IEEE conference on computer vision and pattern recognition},
  pages={3606--3613},
  year={2014}
}

@inproceedings{fei2004learning,
  title={Learning generative visual models from few training examples: An incremental bayesian approach tested on 101 object categories},
  author={Fei-Fei, Li and Fergus, Rob and Perona, Pietro},
  booktitle={2004 conference on computer vision and pattern recognition workshop},
  pages={178--178},
  year={2004},
  organization={IEEE}
}

@inproceedings{parkhi2012cats,
  title={Cats and dogs},
  author={Parkhi, Omkar M and Vedaldi, Andrea and Zisserman, Andrew and Jawahar, CV},
  booktitle={2012 IEEE conference on computer vision and pattern recognition},
  pages={3498--3505},
  year={2012},
  organization={IEEE}
}

@article{maji2013fine,
  title={Fine-grained visual classification of aircraft},
  author={Maji, Subhransu and Rahtu, Esa and Kannala, Juho and Blaschko, Matthew and Vedaldi, Andrea},
  journal={arXiv preprint arXiv:1306.5151},
  year={2013}
}

@inproceedings{nilsback2008automated,
  title={Automated flower classification over a large number of classes},
  author={Nilsback, Maria-Elena and Zisserman, Andrew},
  booktitle={2008 Sixth Indian conference on computer vision, graphics \& image processing},
  pages={722--729},
  year={2008},
  organization={IEEE}
}

@inproceedings{krause20133d,
  title={3d object representations for fine-grained categorization},
  author={Krause, Jonathan and Stark, Michael and Deng, Jia and Fei-Fei, Li},
  booktitle={Proceedings of the IEEE international conference on computer vision workshops},
  pages={554--561},
  year={2013}
}

@inproceedings{singh2022flava,
  title={Flava: A foundational language and vision alignment model},
  author={Singh, Amanpreet and Hu, Ronghang and Goswami, Vedanuj and Couairon, Guillaume and Galuba, Wojciech and Rohrbach, Marcus and Kiela, Douwe},
  booktitle={Proceedings of the IEEE/CVF Conference on Computer Vision and Pattern Recognition},
  pages={15638--15650},
  year={2022}
}

@inproceedings{zhai2022lit,
  title={Lit: Zero-shot transfer with locked-image text tuning},
  author={Zhai, Xiaohua and Wang, Xiao and Mustafa, Basil and Steiner, Andreas and Keysers, Daniel and Kolesnikov, Alexander and Beyer, Lucas},
  booktitle={Proceedings of the IEEE/CVF conference on computer vision and pattern recognition},
  pages={18123--18133},
  year={2022}
}

@inproceedings{yi2023simple,
  title={A simple framework for text-supervised semantic segmentation},
  author={Yi, Muyang and Cui, Quan and Wu, Hao and Yang, Cheng and Yoshie, Osamu and Lu, Hongtao},
  booktitle={Proceedings of the IEEE/CVF Conference on Computer Vision and Pattern Recognition},
  pages={7071--7080},
  year={2023}
}

@inproceedings{du2022learning,
  title={Learning to prompt for open-vocabulary object detection with vision-language model},
  author={Du, Yu and Wei, Fangyun and Zhang, Zihe and Shi, Miaojing and Gao, Yue and Li, Guoqi},
  booktitle={Proceedings of the IEEE/CVF Conference on Computer Vision and Pattern Recognition},
  pages={14084--14093},
  year={2022}
}

@inproceedings{feng2022promptdet,
  title={Promptdet: Towards open-vocabulary detection using uncurated images},
  author={Feng, Chengjian and Zhong, Yujie and Jie, Zequn and Chu, Xiangxiang and Ren, Haibing and Wei, Xiaolin and Xie, Weidi and Ma, Lin},
  booktitle={European Conference on Computer Vision},
  pages={701--717},
  year={2022},
  organization={Springer}
}

@inproceedings{zhong2022regionclip,
  title={Regionclip: Region-based language-image pretraining},
  author={Zhong, Yiwu and Yang, Jianwei and Zhang, Pengchuan and Li, Chunyuan and Codella, Noel and Li, Liunian Harold and Zhou, Luowei and Dai, Xiyang and Yuan, Lu and Li, Yin and others},
  booktitle={Proceedings of the IEEE/CVF conference on computer vision and pattern recognition},
  pages={16793--16803},
  year={2022}
}

@inproceedings{ghiasi2022scaling,
  title={Scaling open-vocabulary image segmentation with image-level labels},
  author={Ghiasi, Golnaz and Gu, Xiuye and Cui, Yin and Lin, Tsung-Yi},
  booktitle={European Conference on Computer Vision},
  pages={540--557},
  year={2022},
  organization={Springer}
}

@inproceedings{
lilanguage,
title={Language-driven Semantic Segmentation},
author={Boyi Li and Kilian Q Weinberger and Serge Belongie and Vladlen Koltun and Rene Ranftl},
booktitle={International Conference on Learning Representations},
year={2022},
url={https://openreview.net/forum?id=RriDjddCLN}
}

@inproceedings{jia2021scaling,
  title={Scaling up visual and vision-language representation learning with noisy text supervision},
  author={Jia, Chao and Yang, Yinfei and Xia, Ye and Chen, Yi-Ting and Parekh, Zarana and Pham, Hieu and Le, Quoc and Sung, Yun-Hsuan and Li, Zhen and Duerig, Tom},
  booktitle={International conference on machine learning},
  pages={4904--4916},
  year={2021},
  organization={PMLR}
}

@article{yuan2021florence,
  title={Florence: A new foundation model for computer vision},
  author={Yuan, Lu and Chen, Dongdong and Chen, Yi-Ling and Codella, Noel and Dai, Xiyang and Gao, Jianfeng and Hu, Houdong and Huang, Xuedong and Li, Boxin and Li, Chunyuan and others},
  journal={arXiv preprint arXiv:2111.11432},
  year={2021}
}

@inproceedings{mahmood2021low,
  title={Low-Budget Active Learning via Wasserstein Distance: An Integer Programming Approach},
  author={Mahmood, Rafid and Fidler, Sanja and Law, Marc T},
  booktitle={International Conference on Learning Representations},
  year={2021}
}

@article{chen2023making,
  title={Making your first choice: to address cold start problem in medical active learning},
  author={Chen, Liangyu and Bai, Yutong and Huang, Siyu and Lu, Yongyi and Wen, Bihan and Yuille, Alan and Zhou, Zongwei},
  journal={Proceedings of Machine Learning Research--nnn},
  volume={1},
  pages={30},
  year={2023}
}

@inproceedings{sinha2019variational,
  title={Variational adversarial active learning},
  author={Sinha, Samarth and Ebrahimi, Sayna and Darrell, Trevor},
  booktitle=ICCV,
  year={2019}
}

@article{yehuda2022active,
  title={Active learning through a covering lens},
  author={Yehuda, Ofer and Dekel, Avihu and Hacohen, Guy and Weinshall, Daphna},
  journal={Advances in Neural Information Processing Systems},
  volume={35},
  pages={22354--22367},
  year={2022}
}

@inproceedings{selvaraju2017gradcam,
  title={Grad-cam: Visual explanations from deep networks via gradient-based localization},
  author={Selvaraju, Ramprasaath R and Cogswell, Michael and Das, Abhishek and Vedantam, Ramakrishna and Parikh, Devi and Batra, Dhruv},
  booktitle={Proceedings of the IEEE international conference on computer vision},
  pages={618--626},
  year={2017}
}

@inproceedings{zhu2023prompt,
  title={Prompt-aligned gradient for prompt tuning},
  author={Zhu, Beier and Niu, Yulei and Han, Yucheng and Wu, Yue and Zhang, Hanwang},
  booktitle={Proceedings of the IEEE/CVF International Conference on Computer Vision},
  pages={15659--15669},
  year={2023}
}

@article{van2008visualizing,
  title={Visualizing data using t-SNE.},
  author={Van der Maaten, Laurens and Hinton, Geoffrey},
  journal={Journal of machine learning research},
  volume={9},
  number={11},
  year={2008}
}

@inproceedings{asghar2016deep,
  title={Deep active learning for dialogue generation},
  author={Asghar, Nabiha and Poupart, Pascal and Jiang, Xin and Li, Hang},
  booktitle = "Proceedings of the 6th Joint Conference on Lexical and Computational Semantics (*{SEM})",
  year={2017}
}

@inproceedings{he2019towards,
  title={Towards better uncertainty sampling: Active learning with multiple views for deep convolutional neural network},
  author={He, Tao and Jin, Xiaoming and Ding, Guiguang and Yi, Lan and Yan, Chenggang},
  booktitle={IEEE International Conference on Multimedia and Expo (ICME)},
  year={2019}
}

@inproceedings{ostapuk2019activelink,
  title={Activelink: deep active learning for link prediction in knowledge graphs},
  author={Ostapuk, Natalia and Yang, Jie and Cudr{\'e}-Mauroux, Philippe},
  booktitle={The World Wide Web Conference (WWW)},
  year={2019}
}

@inproceedings{
fuchsgruber2024uncertainty,
title={Uncertainty for Active Learning on Graphs},
author={Dominik Fuchsgruber and Tom Wollschl{\"a}ger and Bertrand Charpentier and Antonio Oroz and Stephan G{\"u}nnemann},
booktitle={Forty-first International Conference on Machine Learning},
year={2024},
url={https://openreview.net/forum?id=BCEtumPYDt}
}

@inproceedings{ash2019deep,
  title={Deep batch active learning by diverse, uncertain gradient lower bounds},
  author={Ash, Jordan T and Zhang, Chicheng and Krishnamurthy, Akshay and Langford, John and Agarwal, Alekh},
  booktitle=ICLR,
  year={2020}
}

@inproceedings{hwang2022combating,
  title={Combating label distribution shift for active domain adaptation},
  author={Hwang, Sehyun and Lee, Sohyun and Kim, Sungyeon and Ok, Jungseul and Kwak, Suha},
  booktitle=ECCV,
  pages={549--566},
  year={2022},
  organization={Springer}
}

@article{wang2019incorporating,
  title={Incorporating distribution matching into uncertainty for multiple kernel active learning},
  author={Wang, Zengmao and Du, Bo and Tu, Weiping and Zhang, Lefei and Tao, Dacheng},
  journal={IEEE Transactions on Knowledge and Data Engineering (TKDE)},
  year={2019}
}

@inproceedings{hacohen2022active,
  title={Active Learning on a Budget: Opposite Strategies Suit High and Low Budgets},
  author={Hacohen, Guy and Dekel, Avihu and Weinshall, Daphna},
  booktitle={International Conference on Machine Learning},
  pages={8175--8195},
  year={2022},
  organization={PMLR}
}

@inproceedings{NEURIPS2023_2b09bb02,
 author = {Hacohen, Guy and Weinshall, Daphna},
 booktitle = {Advances in Neural Information Processing Systems},
 editor = {A. Oh and T. Naumann and A. Globerson and K. Saenko and M. Hardt and S. Levine},
 pages = {13395--13407},
 publisher = {Curran Associates, Inc.},
 title = {How to Select Which Active Learning Strategy is Best Suited for Your Specific Problem and Budget},
 url = {https://proceedings.neurips.cc/paper_files/paper/2023/file/2b09bb02b90584e2be94ff3ae09289bc-Paper-Conference.pdf},
 volume = {36},
 year = {2023}
}

@InProceedings{Kim_2023_ICCV,
    author    = {Kim, Hoyoung and Oh, Minhyeon and Hwang, Sehyun and Kwak, Suha and Ok, Jungseul},
    title     = {Adaptive Superpixel for Active Learning in Semantic Segmentation},
    booktitle = {Proceedings of the IEEE/CVF International Conference on Computer Vision (ICCV)},
    month     = {October},
    year      = {2023},
    pages     = {943-953}
}

@inproceedings{NEURIPS2023_559a0998,
 author = {Hwang, Sehyun and Lee, Sohyun and Kim, Hoyoung and Oh, Minhyeon and Ok, Jungseul and Kwak, Suha},
 booktitle = {Advances in Neural Information Processing Systems},
 editor = {A. Oh and T. Naumann and A. Globerson and K. Saenko and M. Hardt and S. Levine},
 pages = {27020--27039},
 publisher = {Curran Associates, Inc.},
 title = {Active Learning for Semantic Segmentation with Multi-class Label Query},
 url = {https://proceedings.neurips.cc/paper_files/paper/2023/file/559a0998fab1d19b80e7e43a5852401c-Paper-Conference.pdf},
 volume = {36},
 year = {2023}
}

@inproceedings{
hacohen2023how,
title={How to Select Which Active Learning Strategy is Best Suited for Your Specific Problem and Budget},
author={Guy Hacohen and Daphna Weinshall},
booktitle={Thirty-seventh Conference on Neural Information Processing Systems},
year={2023},
url={https://openreview.net/forum?id=eDDZh8C4W4}
}

@article{
gupte2024revisiting,
title={Revisiting Active Learning in the Era of Vision Foundation Models},
author={Sanket Rajan Gupte and Josiah Aklilu and Jeffrey J Nirschl and Serena Yeung-Levy},
journal={Transactions on Machine Learning Research},
issn={2835-8856},
year={2024},
url={https://openreview.net/forum?id=u8K83M9mbG},
note={}
}

@article{wan2023survey,
  title={A survey of deep active learning for foundation models},
  author={Wan, Tianjiao and Xu, Kele and Yu, Ting and Wang, Xu and Feng, Dawei and Ding, Bo and Wang, Huaimin},
  journal={Intelligent Computing},
  volume={2},
  pages={0058},
  year={2023},
  publisher={AAAS}
}

@inproceedings{kirillov2023segment,
  title={Segment anything},
  author={Kirillov, Alexander and Mintun, Eric and Ravi, Nikhila and Mao, Hanzi and Rolland, Chloe and Gustafson, Laura and Xiao, Tete and Whitehead, Spencer and Berg, Alexander C and Lo, Wan-Yen and others},
  booktitle={Proceedings of the IEEE/CVF International Conference on Computer Vision},
  pages={4015--4026},
  year={2023}
}

@article{achiam2023gpt,
  title={Gpt-4 technical report},
  author={OpenAI},
  journal={arXiv preprint arXiv:2303.08774},
  year={2023}
}

@inproceedings{zhou2022conditional,
  title={Conditional prompt learning for vision-language models},
  author={Zhou, Kaiyang and Yang, Jingkang and Loy, Chen Change and Liu, Ziwei},
  booktitle={Proceedings of the IEEE/CVF conference on computer vision and pattern recognition},
  pages={16816--16825},
  year={2022}
}

@inproceedings{li2023gradient,
  title={Gradient-regulated meta-prompt learning for generalizable vision-language models},
  author={Li, Juncheng and Gao, Minghe and Wei, Longhui and Tang, Siliang and Zhang, Wenqiao and Li, Mengze and Ji, Wei and Tian, Qi and Chua, Tat-Seng and Zhuang, Yueting},
  booktitle={Proceedings of the IEEE/CVF International Conference on Computer Vision},
  pages={2551--2562},
  year={2023}
}

@inproceedings{
dosovitskiy2021an,
title={An Image is Worth 16x16 Words: Transformers for Image Recognition at Scale},
author={Alexey Dosovitskiy and Lucas Beyer and Alexander Kolesnikov and Dirk Weissenborn and Xiaohua Zhai and Thomas Unterthiner and Mostafa Dehghani and Matthias Minderer and Georg Heigold and Sylvain Gelly and Jakob Uszkoreit and Neil Houlsby},
booktitle={International Conference on Learning Representations},
year={2021},
url={https://openreview.net/forum?id=YicbFdNTTy}
}

@article{helber2019eurosat,
  title={Eurosat: A novel dataset and deep learning benchmark for land use and land cover classification},
  author={Helber, Patrick and Bischke, Benjamin and Dengel, Andreas and Borth, Damian},
  journal={IEEE Journal of Selected Topics in Applied Earth Observations and Remote Sensing},
  volume={12},
  number={7},
  pages={2217--2226},
  year={2019},
  publisher={IEEE}
}

@inproceedings{deng2009imagenet,
  title={Imagenet: A large-scale hierarchical image database},
  author={Deng, Jia and Dong, Wei and Socher, Richard and Li, Li-Jia and Li, Kai and Fei-Fei, Li},
  booktitle={2009 IEEE conference on computer vision and pattern recognition},
  pages={248--255},
  year={2009},
  organization={Ieee}
}

@article{emam2021active,
  title={Active learning at the imagenet scale},
  author={Emam, Zeyad Ali Sami and Chu, Hong-Min and Chiang, Ping-Yeh and Czaja, Wojciech and Leapman, Richard and Goldblum, Micah and Goldstein, Tom},
  journal={arXiv preprint arXiv:2111.12880},
  year={2021}
}

@article{codella2019skin,
  title={Skin lesion analysis toward melanoma detection 2018: A challenge hosted by the international skin imaging collaboration (isic)},
  author={Codella, Noel and Rotemberg, Veronica and Tschandl, Philipp and Celebi, M Emre and Dusza, Stephen and Gutman, David and Helba, Brian and Kalloo, Aadi and Liopyris, Konstantinos and Marchetti, Michael and others},
  journal={arXiv preprint arXiv:1902.03368},
  year={2019}
}

@article{tian2020kaokore,
  title={Kaokore: A pre-modern japanese art facial expression dataset},
  author={Tian, Yingtao and Suzuki, Chikahiko and Clanuwat, Tarin and Bober-Irizar, Mikel and Lamb, Alex and Kitamoto, Asanobu},
  journal={arXiv preprint arXiv:2002.08595},
  year={2020}
}

@article{sun2023eva,
  title={Eva-clip: Improved training techniques for clip at scale},
  author={Sun, Quan and Fang, Yuxin and Wu, Ledell and Wang, Xinlong and Cao, Yue},
  journal={arXiv preprint arXiv:2303.15389},
  year={2023}
}
\bibliographystyle{tmlr}

\clearpage
\appendix

\renewcommand{\thesection}{\Alph{section}}
\setcounter{section}{0}

\section{Details of GradFAM}
\label{sec:heatmap_details}

\begin{figure*}[h!]
    \centering
    \begin{minipage}[t]{0.48\linewidth}
        \centering
        \begin{subfigure}[t]{0.31\linewidth}
            \centering
            \setlength{\fboxsep}{0pt}\fbox{\includegraphics[width=\linewidth,height=\linewidth]{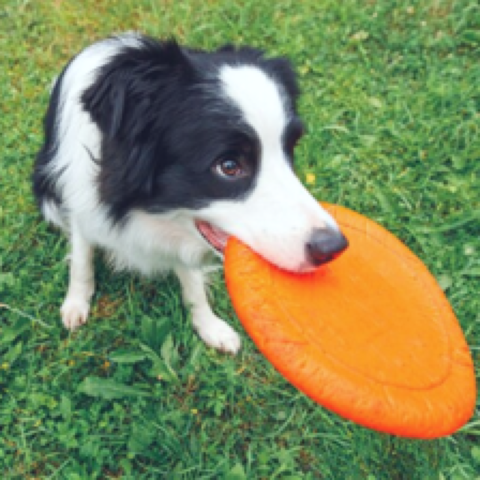}}
            \caption{Image}
        \end{subfigure}\hfill
        \begin{subfigure}[t]{0.31\linewidth}
            \centering
            \setlength{\fboxsep}{0pt}\fbox{\includegraphics[width=\linewidth,height=\linewidth]{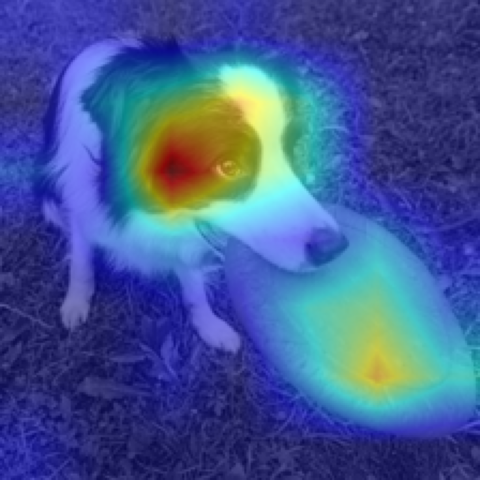}}
            \caption{$I$}
        \end{subfigure}\hfill
        \begin{subfigure}[t]{0.31\linewidth}
            \centering
            \setlength{\fboxsep}{0pt}\fbox{\includegraphics[width=\linewidth,height=\linewidth]{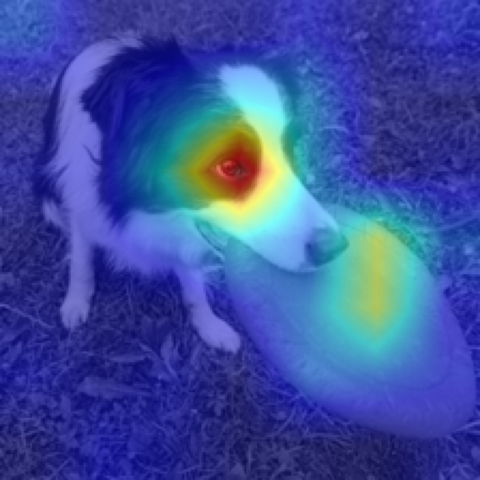}}
            \caption{$\mathcal{F}_{\{\text{Cat},\text{Dog}\}}$}
        \end{subfigure}

        \vspace{1mm}

        \begin{subfigure}[t]{0.31\linewidth}
            \centering
            \setlength{\fboxsep}{0pt}\fbox{\includegraphics[width=\linewidth,height=\linewidth]{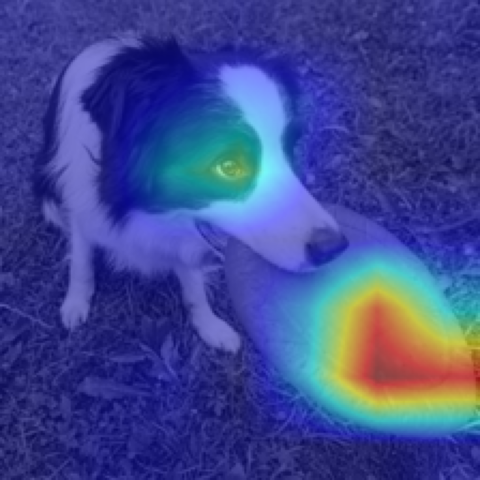}}
            \caption{$\mathcal{F}_{\{\text{Ball},\text{Frisbee}\}}$}
        \end{subfigure}\hfill
        \begin{subfigure}[t]{0.31\linewidth}
            \centering
            \setlength{\fboxsep}{0pt}\fbox{\includegraphics[width=\linewidth,height=\linewidth]{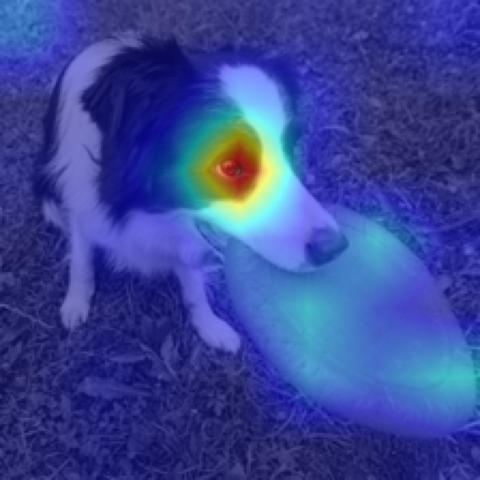}}
            \caption{$\mathcal{F}_{\{\text{Eye},\text{Ring}\}}$}
        \end{subfigure}\hfill
        \begin{subfigure}[t]{0.31\linewidth}
            \centering
            \setlength{\fboxsep}{0pt}\fbox{\includegraphics[width=\linewidth,height=\linewidth]{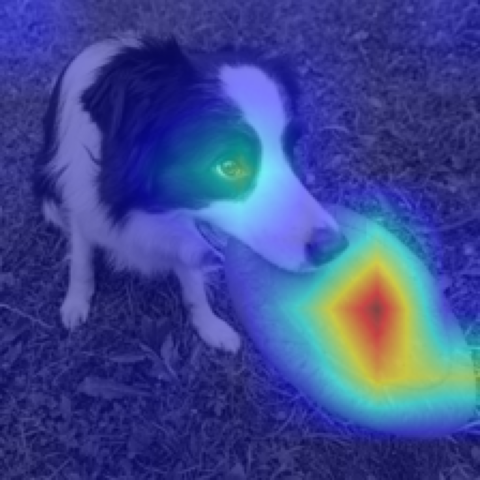}}
            \caption{$\mathcal{F}_{\{\text{Orange},\text{Blue}\}}$}
        \end{subfigure}
    \end{minipage}
    \hfill
    \begin{minipage}[t]{0.48\linewidth}
        \centering
        \begin{subfigure}[t]{0.31\linewidth}
            \centering
            \setlength{\fboxsep}{0pt}\fbox{\includegraphics[width=\linewidth,height=\linewidth]{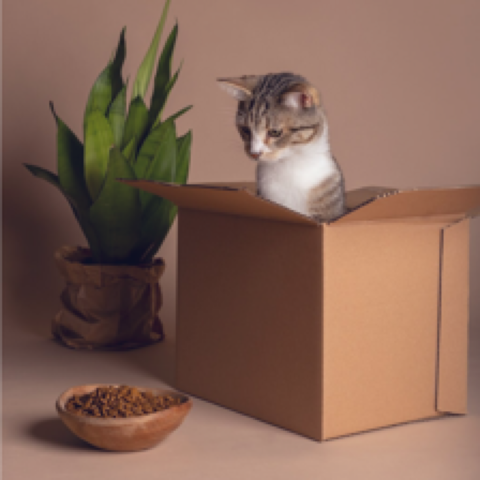}}
            \caption{Image}
        \end{subfigure}\hfill
        \begin{subfigure}[t]{0.31\linewidth}
            \centering
            \setlength{\fboxsep}{0pt}\fbox{\includegraphics[width=\linewidth,height=\linewidth]{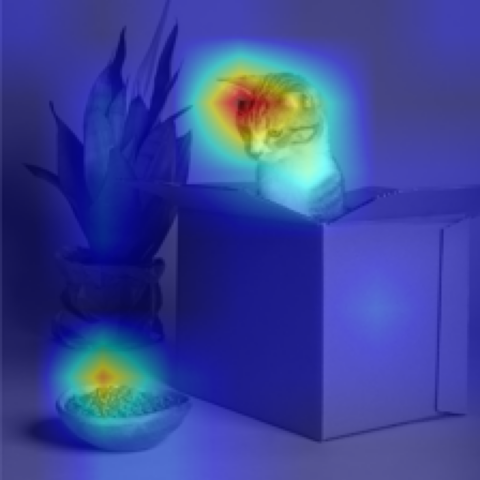}}
            \caption{$I$}
        \end{subfigure}\hfill
        \begin{subfigure}[t]{0.31\linewidth}
            \centering
            \setlength{\fboxsep}{0pt}\fbox{\includegraphics[width=\linewidth,height=\linewidth]{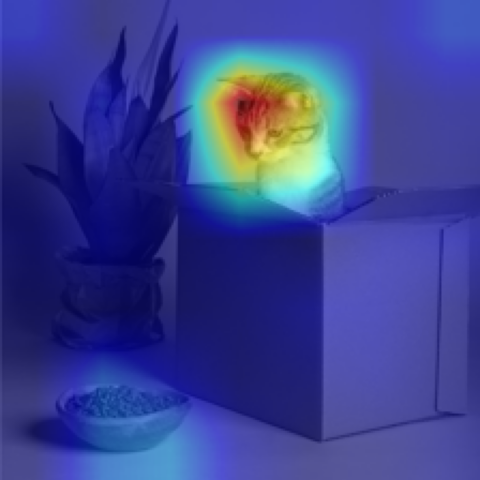}}
            \caption{$\mathcal{F}_{\{\text{Cat},\text{Dog}\}}$}
        \end{subfigure}

        \vspace{1mm}

        \begin{subfigure}[t]{0.31\linewidth}
            \centering
            \setlength{\fboxsep}{0pt}\fbox{\includegraphics[width=\linewidth,height=\linewidth]{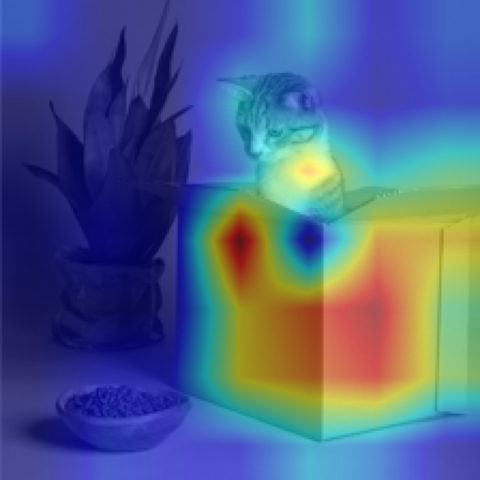}}
            \caption{$\mathcal{F}_{\{\text{Ball},\text{Box}\}}$}
        \end{subfigure}\hfill
        \begin{subfigure}[t]{0.31\linewidth}
            \centering
            \setlength{\fboxsep}{0pt}\fbox{\includegraphics[width=\linewidth,height=\linewidth]{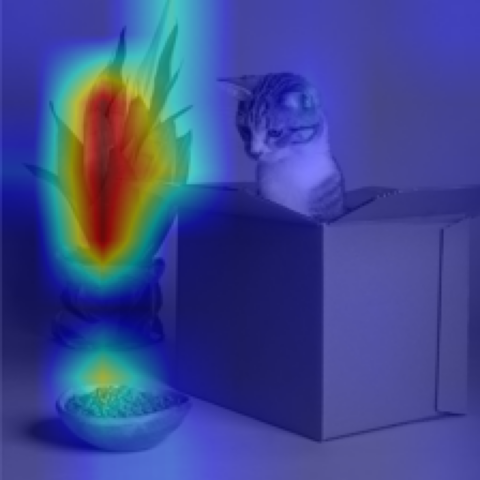}}
            \caption{$\mathcal{F}_{\{\text{Plant},\text{Water}\}}$}
        \end{subfigure}\hfill
        \begin{subfigure}[t]{0.31\linewidth}
            \centering
            \setlength{\fboxsep}{0pt}\fbox{\includegraphics[width=\linewidth,height=\linewidth]{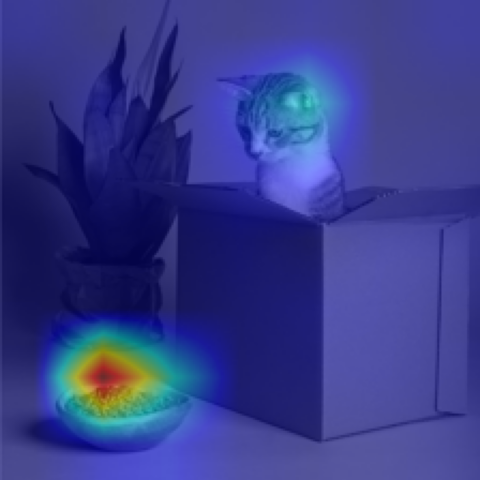}}
            \caption{$\mathcal{F}_{\{\text{Feed},\text{Toy}\}}$}
        \end{subfigure}
    \end{minipage}

    \caption{GradFAM visualizations for two image examples under diverse class-guided feature targets.}
    \label{fig:heatmap-combined}
\end{figure*}

In Section~\ref{sec:multimodal-clustering}, we propose class-guided features and analyze them with our GradFAM, a modified version of GradCAM~\citep{selvaraju2017gradcam}.
Here, we provide a detailed description of GradFAM and its differences from GradCAM.

\vspace{1mm}
\noindent\textbf{Gradient-weighted class activation map (CAM).}
To highlight the importance of regions in an image $x$ associated with a given class $c$, Grad-CAM introduces the class-discriminative localization map as follows:
\begin{align}
L^c_{\text{CAM}}(x) \in \mathbb{R}^{U \times V} \;,
\end{align}
where $U$ and $V$ are the width and height of the image $x$.

Let $f(x; \theta) \in \mathbb{R}^{|\mathcal{C}|}$ represent the output logits of a neural network with parameters $\theta$, where $|\mathcal{C}|$ is the total number of classes. For a given class $c \in \mathcal{C}$, the score $y^c(x; \theta)$ is defined as follows:
\begin{align}
y^c(x; \theta) := f_c(x; \theta) \;,
\end{align}
where $f_c(x; \theta)$ denotes the $c$-th logit value, representing the model's confidence for class $c$ given the image $x$.
Grad-CAM computes the gradient of $y^c(x; \theta)$ with respect to the activation maps $A^k$ in the last convolutional layer, where $A^k \in \mathbb{R}^{W \times H}$ represents the activation map of the $k$-th channel with spatial dimensions $W$ and $H$. Note that $W$ and $H$ are typically smaller than $U$ and $V$ due to downsampling in the convolutional layers.

The importance weight $\alpha_c^k$ for each channel $k$ is obtained by applying global average pooling over the gradients as:
\begin{align}
\alpha_{\text{CAM}}^{c,k} := \frac{1}{W \times H} \sum_{w=1}^W \sum_{h=1}^H \frac{\partial y^c(x; \theta)}{\partial A_{w,h}^k} \;,
\label{eq:gradcam-weights}
\end{align}
representing the overall importance of the $k$-th feature map for predicting class $c$.

The class-discriminative localization map $L^c_{\text{CAM}}(x)$ is then computed as a weighted combination of the activation maps, followed by upsampling to match the dimensions of the input image $x$:
\begin{align}
L^c_{\text{CAM}}(x) := \mathcal{U} \left( \text{ReLU}\left(\sum_k \alpha_{\text{CAM}}^{c,k} A^k \right) \right) \;,
\end{align}
where $\mathcal{U}$ denotes the upsampling function.

\input{Figures_tex_sup/fig_sup_full_iccv}

\begin{figure*}[t!]
    \begin{subfigure}[t]{0.24\textwidth}
        \centering
        \begin{tikzpicture}
            \node[anchor=north west, inner sep=0] (image) at (0, 0) {\fbox{\includegraphics[width=0.8\linewidth, height=0.8\linewidth]{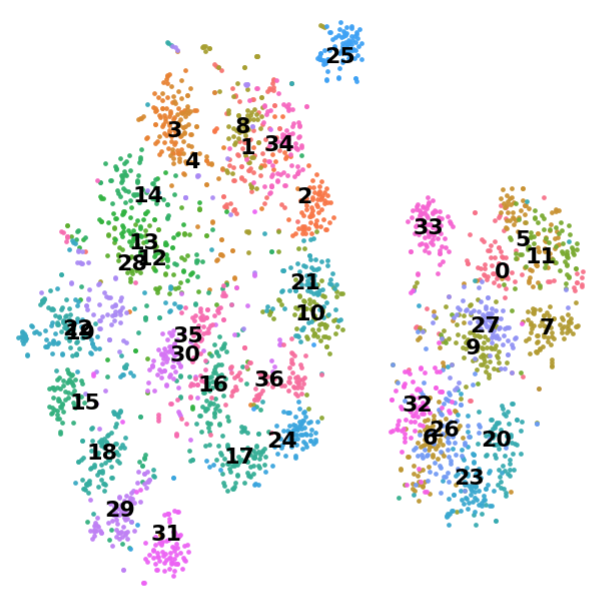}}};
            \node[anchor=north west] at (image.north west) {\tiny ARI: 48.00\%};
        \end{tikzpicture}
        \caption{Image features}
    \end{subfigure} %
    \hfill
    \begin{subfigure}[t]{0.24\textwidth}
        \centering
        \begin{tikzpicture}
            \node[anchor=north west, inner sep=0] (image) at (0, 0) {\fbox{\includegraphics[width=0.8\linewidth, height=0.8\linewidth]{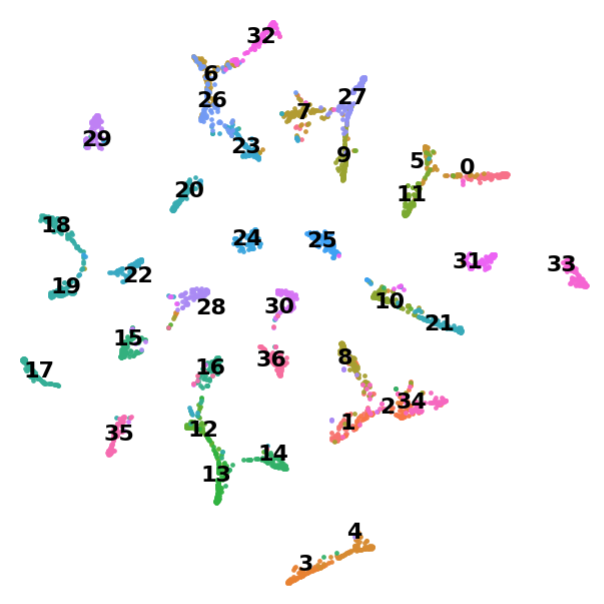}}};
            \node[anchor=north west] at (image.north west) {\tiny ARI: 75.34\%};
        \end{tikzpicture}
        \caption{Weighted text features}
    \end{subfigure} %
    \hfill
    \begin{subfigure}[t]{0.24\textwidth}
        \centering
        \begin{tikzpicture}
            \node[anchor=north west, inner sep=0] (image) at (0, 0) {\fbox{\includegraphics[width=0.8\linewidth, height=0.8\linewidth]{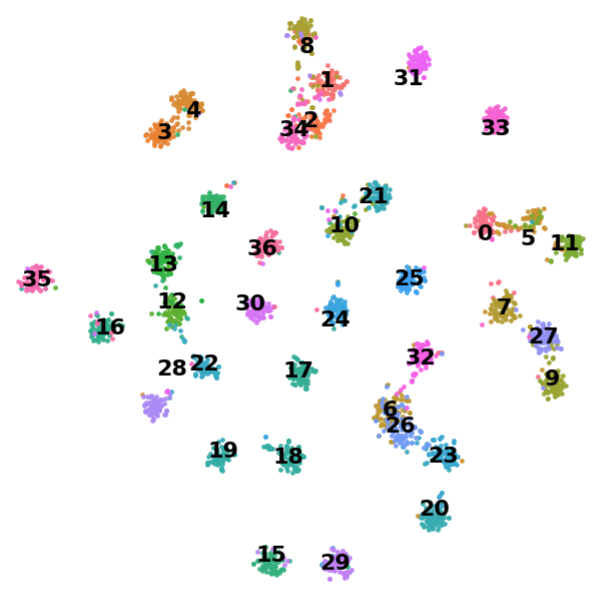}}};
            \node[anchor=north west] at (image.north west) {\tiny ARI: 77.60\%};
        \end{tikzpicture}
        \caption{Class-guided features with pseudo-labels}
    \end{subfigure} %
    \hfill
    \begin{subfigure}[t]{0.24\textwidth}
        \centering
        \begin{tikzpicture}
            \node[anchor=north west, inner sep=0] (image) at (0, 0) {\fbox{\includegraphics[width=0.8\linewidth, height=0.8\linewidth]{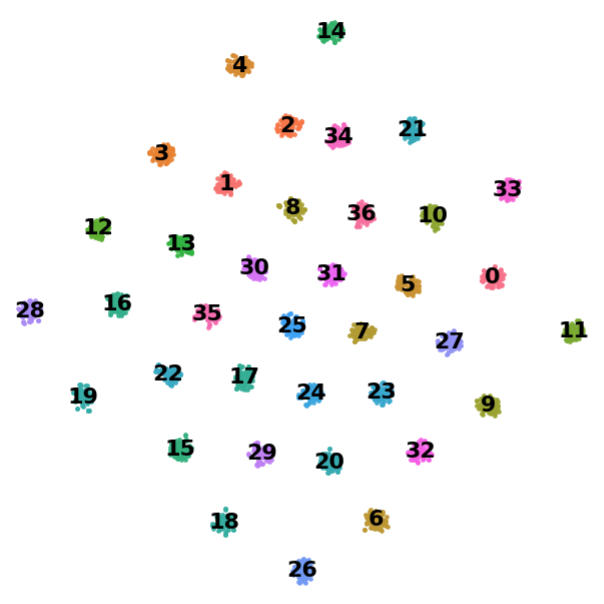}}};
            \node[anchor=north west] at (image.north west) {\tiny ARI: 100\%};
        \end{tikzpicture}
        \caption{Class-guided features with ground-truths}
        \label{fig:sup-tsne-pets-d}
    \end{subfigure}
    \caption{{\em Various clustering methods on OxfordPets dataset.} (a) Clustering based solely on image features primarily separates the data into two large groups, corresponding to dogs and cats, but fails to capture finer details. (b) Clustering on weighted text features improves Adjusted Rand Index (ARI), but some clusters remain ambiguous. (c) Class-guided clustering with pseudo-labels produces more distinct clusters, aligned with the guided class set $|\mathcal{C}| = 37$. (d) Class-guided clustering with ground truth shows perfect alignment of clusters.}
    \label{fig:sup-tsne-pets}
\end{figure*}

\noindent\textbf{Gradient-weighted feature activation map (FAM).}
To adapt GradCAM for vision-language models (VLMs), such as CLIP, which consist of an image encoder $\theta_{\text{img}}$ and a text encoder $\theta_{\text{txt}}$, we redefine the weight $\alpha$ with target features $\mathcal{F}_{\text{target}}$ as follows:
\begin{align}
\alpha_{\text{FAM}}^{k} := \frac{1}{W \times H} \sum_{w=1}^W \sum_{h=1}^H \frac{\partial 
\text{cos}(\theta_{\text{img}}(x), \mathcal{F}_{\text{target}})}{\partial A_{w,h}^k} \;,
\end{align}
representing the overall important of the $k$-th feature map in determining the cosine similarity between the image features $\theta_{\text{img}}(x)$ and the target features $\mathcal{F}_{\text{target}}$.
Here, the key difference from the GradCAM's weights in~\eqref{eq:gradcam-weights} lies in the absence of a specific class $c$.
This modification enables the use of GradFAM for more flexible and label-independent analyses, accommodating the multimodal nature of VLMs.
Building on this, the target feature discriminative localization map $L_{\text{FAM}}(x)$ is then computed as:
\begin{align}
L_{\text{FAM}}(x) := \mathcal{U} \left( \text{ReLU}\left(\sum_k \alpha_{\text{FAM}}^{k} A^k \right) \right) \;.
\end{align}

\input{Figures_tex_sup/fig_sup_harmonic}

Our GradFAM can visualize the importance of various target features on the image, including (i) $\mathcal{F}_{\text{target}} = \theta_{\text{txt}}(c)$, (ii) $\mathcal{F}_{\text{target}} = \theta_{\text{img}}(x)$, and (iii) $\mathcal{F}_{\text{target}} = \mathcal{F}_{\mathcal{C}}(x)$. 
Specifically, when $\mathcal{F}_{\text{target}} = \theta_{\text{txt}}(c)$, our GradFAM operates almost identically to the original GradCAM.
For $\mathcal{F}_{\text{target}} = \theta_{\text{img}}(x)$, due to VLMs being trained through contrastive learning to align images and texts in a shared embedding space, all objects in the image are highlighted.
In the case of our class-guided features $\mathcal{F}_{\mathcal{C}}(x)$ in~\eqref{eq:multimodal-embeddings}, the guiding class set $\mathcal{C}$ effectively highlights the corresponding objects.

\begin{figure*}[t!]
    \centering
    \begin{subfigure}[t]{0.28\linewidth}
        \centering
        \begin{tikzpicture}
            \node[anchor=north west, inner sep=0] (image) at (0, 0) {\fbox{\includegraphics[width=0.8\linewidth, height=0.8\linewidth]{Figures/tsne_bird_i_2.png}}};
            \node[anchor=north west] at (image.north west) {\tiny ARI: 45.27\%};
        \end{tikzpicture}
        \caption{\scriptsize Image features}
    \end{subfigure} %
    \hspace{5mm}
    \begin{subfigure}[t]{0.28\linewidth}
        \centering
        \begin{tikzpicture}
            \node[anchor=north west, inner sep=0] (image) at (0, 0) {\fbox{\includegraphics[width=0.8\linewidth, height=0.8\linewidth]{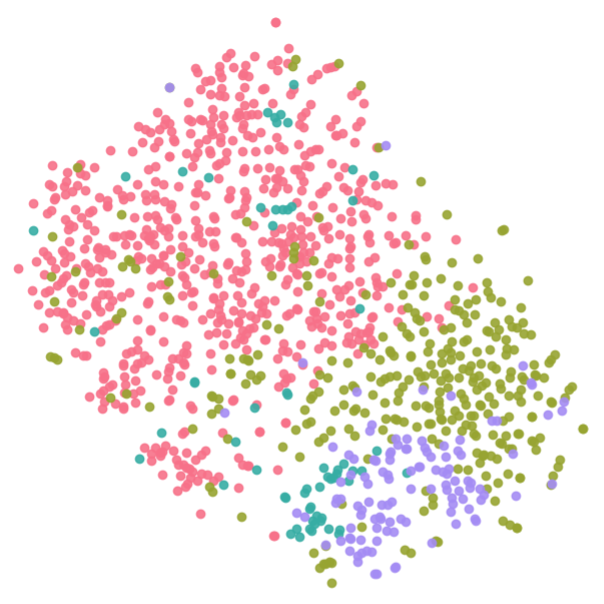}}};
            \node[anchor=north west] at (image.north west) {\tiny ARI: 22.30\%};
        \end{tikzpicture}
        \caption{\scriptsize Image features}
    \end{subfigure} %
    \hspace{5mm}
    \begin{subfigure}[t]{0.28\linewidth}
        \centering
        \begin{tikzpicture}
            \node[anchor=north west, inner sep=0] (image) at (0, 0) {\fbox{\includegraphics[width=0.8\linewidth, height=0.8\linewidth]{Figures/tsne_bird_i_20.png}}};
            \node[anchor=north west] at (image.north west) {\tiny ARI: 12.74\%};
        \end{tikzpicture}
        \caption{\scriptsize Image features}
    \end{subfigure}
    \vspace{2mm}
    \\
    \begin{subfigure}[t]{0.28\linewidth}
        \centering
        \begin{tikzpicture}
            \node[anchor=north west, inner sep=0] (image) at (0, 0) {\fbox{\includegraphics[width=0.8\linewidth, height=0.8\linewidth]{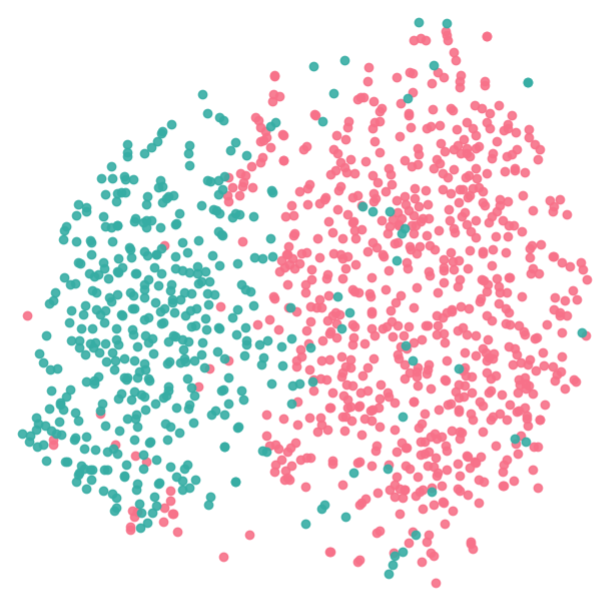}}};
            \node[anchor=north west] at (image.north west) {\tiny ARI: 57.33\%};
        \end{tikzpicture}
        \caption{\scriptsize Class-guided features, $|\mathcal{C}| =2$}
    \end{subfigure} %
    \hspace{5mm}
    \begin{subfigure}[t]{0.28\linewidth}
        \centering
        \begin{tikzpicture}
            \node[anchor=north west, inner sep=0] (image) at (0, 0) {\fbox{\includegraphics[width=0.8\linewidth, height=0.8\linewidth]{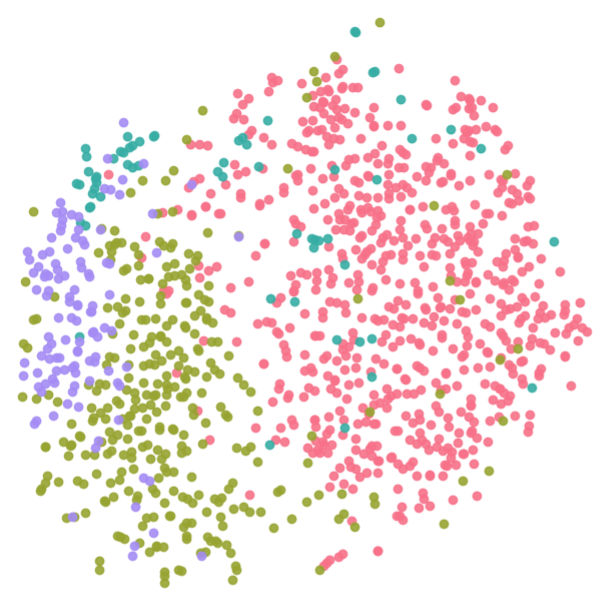}}};
            \node[anchor=north west] at (image.north west) {\tiny ARI: 28.01\%};
        \end{tikzpicture}
        \caption{\scriptsize Class-guided features, $|\mathcal{C}| =4$}
    \end{subfigure} %
    \hspace{5mm}
    \begin{subfigure}[t]{0.28\linewidth}
        \centering
        \begin{tikzpicture}
            \node[anchor=north west, inner sep=0] (image) at (0, 0) {\fbox{\includegraphics[width=0.8\linewidth, height=0.8\linewidth]{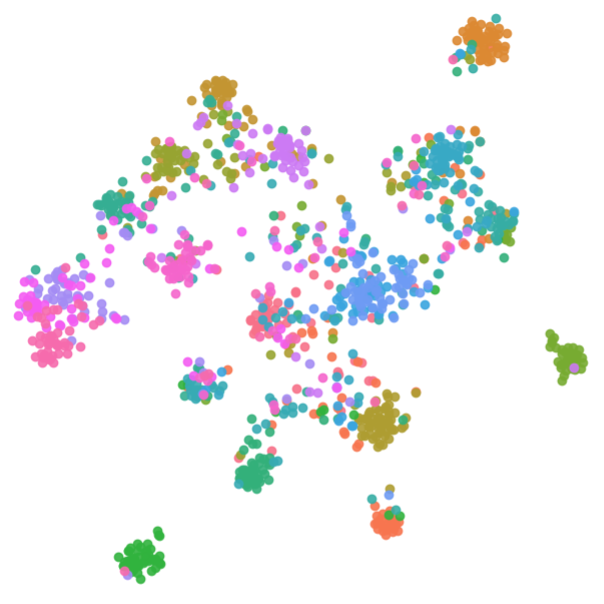}}};
            \node[anchor=north west] at (image.north west) {\tiny ARI: 39.68\%};
        \end{tikzpicture}
        \caption{\scriptsize Class-guided features, $|\mathcal{C}| =20$}
    \end{subfigure}
    \caption{{\em Class-guided clustering on WaterBirds dataset.} (a, b, c) Clustering based solely on image features results in clusters that are poorly separated. (d, e, f) In contrast, our class-guided clustering, which incorporates class information, leads to more distinct clusters that align with the size of the guiding  class set $\mathcal{C}$.}
    \label{fig:sup-tsne-birds}
\end{figure*}

\begin{figure*}[t!]
    \begin{subfigure}[t]{0.24\textwidth}
        \centering
        \begin{tikzpicture}
            \node[anchor=north west, inner sep=0] (image) at (0, 0) {\fbox{\includegraphics[width=0.8\linewidth, height=0.8\linewidth]{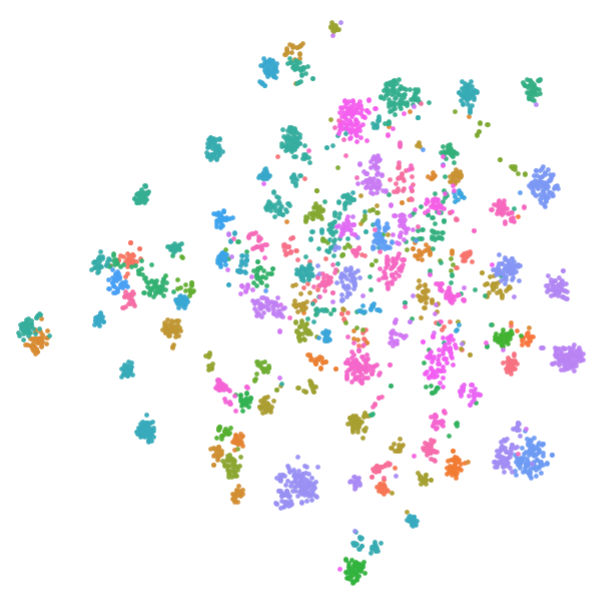}}};
            \node[anchor=north west] at (image.north west) {\tiny ARI: 67.69\%};
        \end{tikzpicture}
        \caption{\scriptsize Image features, $r=0$}
    \end{subfigure} %
    \hfill
    \begin{subfigure}[t]{0.24\textwidth}
        \centering
        \begin{tikzpicture}
            \node[anchor=north west, inner sep=0] (image) at (0, 0) {\fbox{\includegraphics[width=0.8\linewidth, height=0.8\linewidth]{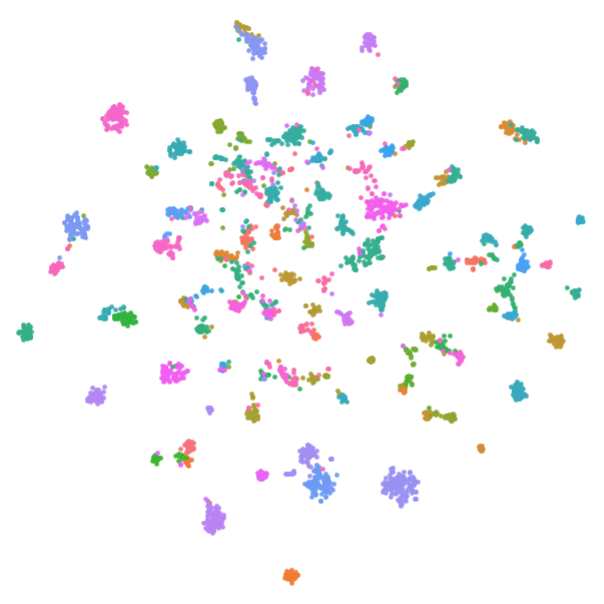}}};
            \node[anchor=north west] at (image.north west) {\tiny ARI: 70.62\%};
        \end{tikzpicture}
        \caption{\scriptsize Class-guided features, $r=0$}
    \end{subfigure} %
    \hfill
    \begin{subfigure}[t]{0.24\textwidth}
        \centering
        \begin{tikzpicture}
            \node[anchor=north west, inner sep=0] (image) at (0, 0) {\fbox{\includegraphics[width=0.8\linewidth, height=0.8\linewidth]{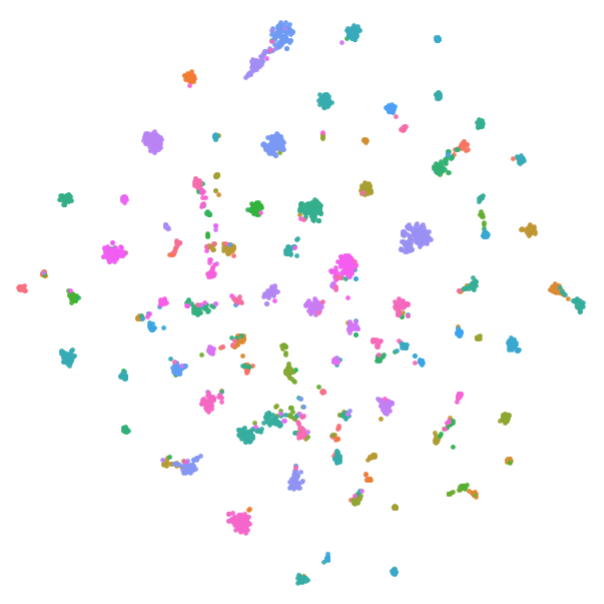}}};
            \node[anchor=north west] at (image.north west) {\tiny ARI: 79.01\%};
        \end{tikzpicture}
        \caption{\scriptsize Class-guided features, $r=4$}
    \end{subfigure} %
    \hfill
    \begin{subfigure}[t]{0.24\textwidth}
        \centering
        \begin{tikzpicture}
            \node[anchor=north west, inner sep=0] (image) at (0, 0) {\fbox{\includegraphics[width=0.8\linewidth, height=0.8\linewidth]{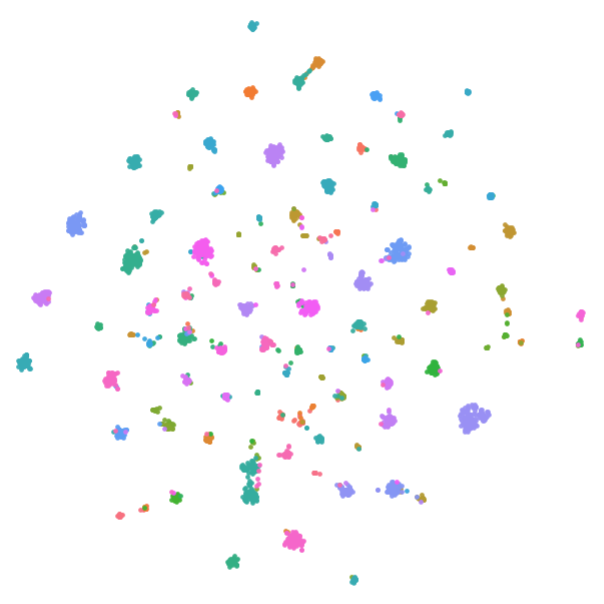}}};
            \node[anchor=north west] at (image.north west) {\tiny ARI: 84.17\%};
        \end{tikzpicture}
        \caption{\scriptsize Class-guided features, $r=8$}
    \end{subfigure}
    \caption{{\em Class-guided clustering across different rounds.} (a, b) In the initial round, \ie  $r=0$, clustering on class-guided features achieves a higher ARI compared to clustering on image features. (c, d) We observe an improvement in ARI performance as the rounds progress.}
    \label{fig:sup-tsne-rounds}
\end{figure*}


\section{Active Learning with Conserved Budgets}
\label{sec:additional-experiments}

For fair comparisons, we evaluate our~\sftype{CB+SQ} alongside with baselines over eight rounds, as illustrated in Figure~\ref{fig:main-graph}.
However, thanks to our selective querying in Section~\ref{sec:budget-saving}, we conserve budgets for each round.
Here, we conduct additional experiments using the conserved budget.
Figure~\ref{fig:sup-full} shows that our~\sftype{CB+SQ} outperforms the baselines under various budget scenarios.

\begin{table*}[t!]
\caption{{\em Examples of guiding class set.} To improve the clarity of clustering visualization, we randomly subsample 20 classes from the total of 200 classes.}
\label{tab:waterbirds-classes-20}
\centering
\footnotesize
\begin{adjustbox}{max width=\textwidth}
\begin{tabular}{clclclcl}
\toprule
\textbf{Index} & \textbf{Name} & \textbf{Index} & \textbf{Name} & \textbf{Index} & \textbf{Name} & \textbf{Index} & \textbf{Name} \\
\midrule
019 & Gray Catbird & 027 & Shiny Cowbird & 029 & American Crow & 032 & Mangrove Cuckoo \\
046 & Gadwall & 057 & Rose-breasted Grosbeak & 058 & Pigeon Guillemot & 061 & Heermann's Gull \\
069 & Rufous Hummingbird & 073 & Blue Jay & 079 & Belted Kingfisher & 098 & Scott Oriole \\
116 & Chipping Sparrow & 121 & Grasshopper Sparrow & 125 & Lincoln Sparrow & 155 & Warbling Vireo \\
167 & Hooded Warbler & 172 & Nashville Warbler & 174 & Palm Warbler & 185 & Bohemian Waxwing \\
\bottomrule
\end{tabular}
\end{adjustbox}
\end{table*}

\section{Analyses of Class-Guided Clustering}
\label{sup:class-guided-clustering}
In Figure~\ref{fig:tsne-birds}, we visualize the difference between class-guided clustering and conventional clustering based on image features.
In this section, we present additional analyses on various clustering methods across different datasets and rounds.

\vspace{1mm}
\noindent\textbf{Class-guided clustering on OxfordPets dataset.}
In Figure~\ref{fig:sup-tsne-pets}, we analyze various clustering methods, including clustering on (a) image features in~\eqref{eq:image-features}, (b) weighted text features in~\eqref{eq:weighted-text}, (c) class-guided features with pseudo-labels in~\eqref{eq:multimodal-embeddings}, (d) class-guided features with ground-truth labels, where the weights of weighted text features are replaced to ground-truth labels, \ie one-hot encodings on labels.
Figure~\ref{fig:sup-tsne-pets} illustrates that clustering on class-guided features achieves higher Adjusted Rand Index (ARI) values.
Especially, Figure~\ref{fig:sup-tsne-pets-d} suggests that as the performance of VLMs improves, perfect clustering becomes achievable.

\begin{table}[!t]
\caption{{\em Various guiding class sets.} Based on two habitats and two backgrounds, we construct class sets comprising 2 and 4 classes.}
\label{tab:waterbirds-classes-2}
\centering
\footnotesize
\begin{tabular}{@{}cl@{}}
\toprule
\textbf{\( |\mathcal{C}| \)} & $\mathcal{C}$ \\ \midrule
2 & 
\begin{tabular}[c]{@{}l@{}}
land background \\
water background
\end{tabular} \\ \midrule
4 & 
\begin{tabular}[c]{@{}l@{}}
land bird on land background \\
land bird on water background \\
water bird on land background \\
water bird on water background
\end{tabular} \\ \bottomrule
\end{tabular}
\end{table}

\begin{table*}[t!]
\begin{center}
\caption{{\em Various values of $K$ for $K$-means clustering.} Our progressively increasing $K$ across rounds achieves the best performance throughtout all rounds.}
\label{tab:various-k}
\setlength\tabcolsep{5pt}
\centering
\footnotesize
\begin{tabular}{ccccccccc}
\toprule
\textbf{Methods} & \textbf{Round 1} & \textbf{Round 2} & \textbf{Round 3} & \textbf{Round 4} & \textbf{Round 5} & \textbf{Round 6} & \textbf{Round 7} & \textbf{Round 8} \\ \midrule
$K = B$ & 63.65$_{\pm 0.16}$ & 66.37$_{\pm 0.16}$ & 67.07$_{\pm 0.49}$ & 68.34$_{\pm 0.46}$ & 69.15$_{\pm 0.16}$ & 69.43$_{\pm 0.36}$ & 69.96$_{\pm 0.10}$ & 70.28$_{\pm 0.21}$ \\
$K = 8 \times B$ & 59.77$_{\pm 0.27}$ & 64.30$_{\pm 0.59}$ & 67.33$_{\pm 0.57}$ & 69.30$_{\pm 0.48}$ & 70.62$_{\pm 0.63}$ & 71.92$_{\pm 0.54}$ & 72.46$_{\pm 0.32}$ & 72.96$_{\pm 0.26}$ \\
\rowcolor{Gray}
$K = r \times B$ & \textbf{63.85}$_{\pm 0.26}$ & \textbf{66.91}$_{\pm 0.38}$ & \textbf{68.92}$_{\pm 0.81}$ & \textbf{70.49}$_{\pm 0.06}$ & \textbf{71.21}$_{\pm 0.18}$ & \textbf{72.16}$_{\pm 0.28}$ & \textbf{72.94}$_{\pm 0.09}$ & \textbf{73.34}$_{\pm 0.12}$ \\
\bottomrule
\end{tabular}
\end{center}
\end{table*}

\vspace{1mm}
\noindent\textbf{Class-guided clustering on WaterBirds dataset.}
\label{sec:waterbirds-details}
In Section~\ref{sec:further-analyses} and Figure~\ref{fig:tsne-birds}, we introduce the WaterBirds dataset to demonstrate the benefits of class-guided clustering, where different classes are guided within the same dataset.
Specifically, the Waterbirds dataset comprises 200 distinct bird species, with each image annotated by habitat (water, land), background (water, land), and specific species.
For our analyses, we select 20 classes and leverage various label information to separate the subset into groups of 2, 4, and 20 classes.
Tables~\ref{tab:waterbirds-classes-20} and~\ref{tab:waterbirds-classes-2} provide the detailed class names.
We note that text prompts such as “a photo of a ” are prepended to each class $c \in \mathcal{C}$ to generate final prompts.
As shown in Figure~\ref{fig:sup-tsne-birds}, class-guided features based on different sizes of guiding class sets $\mathcal{C}$ effectively represent class-specific information.

\vspace{2mm}
\noindent\textbf{Cluster-Guided Clustering with Various Rounds.}
We analyze the impact of class-guided clustering, derived from the zero-shot CLIP model, in comparison to various other clustering methods before initiating active learning. 
Here, we investigate cluster-guided clustering on the OxfordFlowers dataset, utilizing text prompts that evolve with each round. 
Figure~\ref{fig:sup-tsne-rounds} illustrates that as the rounds progress, class-guided clustering forms increasingly well-separated clusters, accompanied by a steady increase in ARI.

\begin{table}[!t]
\centering
\footnotesize
\caption{\em{Experiments on the ISIC dataset.}}
\label{tab:isic}
\begin{tabular}{c|ccccc}
\toprule
\textbf{Budget (\%)} & \textbf{Entropy} & \textbf{CoreSet} & \textbf{BADGE} & \textbf{PCB} & \textbf{CB (ours)} \\
\midrule
25  & 57.87 & 53.69 & 58.37 & 51.49 & \textbf{63.38} \\
50  & 61.35 & 59.21 & 62.10 & 57.52 & \textbf{64.01} \\
75  & 61.65 & 60.81 & 63.89 & 62.10 & \textbf{64.41} \\
100 & 62.25 & 60.91 & 65.59 & 64.54 & \textbf{65.70} \\
\bottomrule
\end{tabular}
\end{table}

\begin{table}[!t]
\centering
\footnotesize
\caption{\em{Experiments on the KaoKore dataset.}}
\label{tab:kaokore}
\begin{tabular}{c|ccccc}
\toprule
\textbf{Budget (\%)} & \textbf{Entropy} & \textbf{CoreSet} & \textbf{BADGE} & \textbf{PCB} & \textbf{CB (ours)} \\
\midrule
25  & 51.68 & 45.75 & 45.36 & 50.74 & \textbf{52.62} \\
50  & 56.36 & 53.32 & 55.58 & 56.52 & \textbf{56.83} \\
75  & 57.73 & 57.06 & 59.02 & 57.30 & \textbf{59.33} \\
100 & 60.11 & 59.25 & 61.01 & 60.58 & \textbf{61.90} \\
\bottomrule
\end{tabular}
\end{table}

\begin{table}[!t]
\centering
\footnotesize
\caption{\em{Experiments on the DTD dataset with CLIP ViT-L/14-336.}}
\label{tab:dtd_l14}
\begin{tabular}{c|ccccc}
\toprule
\textbf{Budget (\%)} & \textbf{Entropy} & \textbf{CoreSet} & \textbf{BADGE} & \textbf{PCB} & \textbf{CB (ours)} \\
\midrule
25  & 43.00 & 41.13 & 44.25 & 50.49 & \textbf{64.10} \\
50  & 57.60 & 52.66 & 60.74 & 64.44 & \textbf{68.83} \\
75  & 65.33 & 61.70 & 66.19 & 68.93 & \textbf{70.90} \\
100 & 69.41 & 65.17 & 69.62 & 72.60 & \textbf{72.97} \\
\bottomrule
\end{tabular}
\end{table}

\begin{table}[!t]
\centering
\footnotesize
\caption{\em{Experiments on the DTD dataset with EVA01-CLIP-g-14-plus (ViT-H/14).}}
\label{tab:dtd_eva}
\begin{tabular}{c|ccccc}
\toprule
\textbf{Budget (\%)} & \textbf{Entropy} & \textbf{CoreSet} & \textbf{BADGE} & \textbf{PCB} & \textbf{CB (ours)} \\
\midrule
25  & 40.17 & 39.38 & 44.46 & 43.87 & \textbf{58.27} \\
50  & 58.45 & 52.44 & 60.09 & 63.95 & \textbf{68.32} \\
75  & 67.18 & 60.72 & 67.65 & 69.11 & \textbf{71.81} \\
100 & 71.61 & 66.77 & 71.91 & 72.71 & \textbf{73.92} \\
\bottomrule
\end{tabular}
\end{table}

\vspace{1mm}
\noindent\textbf{Effect of Various $K$.}
\label{app:kmeans}
In Section~\ref{sec:acquisition}, we set $K$ equal to the budget $B$ in the initial round and introduce a linearly increasing $K$ according to round $r$, \ie $K = r \times B$.
Here, we analyze the effect of this increasing $K$.
Table~\ref{tab:various-k} shows that fixed values of $K$, whether small $(K = B)$ or large $(K = 8 \times B)$, are less effective compared to our incrementally increasing $K$.
Specifically, using a small $K$ results in multiple samples being selected from the same cluster, leading to redundancy and reduced performance.
On the other hand, a large $K$ fails to select representative samples during the initial round, resulting in diminished performance.

\section{Additional Ablation Studies}
\noindent\textbf{Extension to Non-Natural Image Domains.}
We conduct additional experiments on the medical dataset ISIC~\citep{codella2019skin} in Table~\ref{tab:isic} and the illustrative dataset KaoKore~\citep{tian2020kaokore} in Table~\ref{tab:kaokore}. While the improvements are less pronounced than those on our main benchmarks in Figure~\ref{fig:main-graph}, the proposed class-balanced (CB) acquisition still consistently outperforms strong baselines, indicating that the approach remains effective even when VLM priors are less directly aligned with the target domain.

\vspace{1mm}
\noindent\textbf{Generalization to Other Backbones.}
We utilize CLIP ViT-B/32 architecture for main experiments. Here, we extend the evaluation to larger backbones: CLIP ViT-L/14-336 and EVA01-CLIP-g-14-plus (ViT-H/14)~\citep{sun2023eva}. Tables~\ref{tab:dtd_l14} and \ref{tab:dtd_eva} show results on the DTD dataset. Here, we follows the same experimental setup with Figure~\ref{fig:main-graph}, except for the backbone. Increasing model capacity generally improves the final performance when the annotation budget is fully consumed, and CB remains consistently superior to baselines, supporting the generalizability of our acquisition strategy across VLM backbones.

\begin{table}[t]
\centering
\footnotesize
\caption{\em{Effect of class-guided features on initial pool selection.} In the initial round, incorporating class-guided features consistently improves baseline performance across six datasets. * indicates that K-means clustering is employed.}
\label{tab:init_class_guided}
\setlength{\tabcolsep}{4pt}
\renewcommand{\arraystretch}{1.05}
\resizebox{\textwidth}{!}{
\begin{tabular}{ll|ccccccc}
\toprule
\textbf{Type} & \textbf{Method} & \textbf{Caltech101} & \textbf{OxfordPets} & \textbf{StanfordCars} & \textbf{Flowers102} & \textbf{FGVCAircraft} & \textbf{DTD} & \textbf{Avg} \\
\midrule
-- & Random & $62.10_{\pm 0.42}$ & $46.95_{\pm 0.88}$ & $30.56_{\pm 0.54}$ & $53.36_{\pm 3.01}$ & $14.64_{\pm 0.37}$ & $25.35_{\pm 1.98}$ & $38.83_{\pm 0.65}$ \\
\midrule
Image Features & TypiClust & $77.05_{\pm 0.20}$ & $58.27_{\pm 0.17}$ & $36.53_{\pm 0.15}$ & $76.64_{\pm 0.16}$ & $17.46_{\pm 0.49}$ & $41.11_{\pm 0.28}$ & $51.18_{\pm 0.11}$ \\
Image Features & ProbCover & $51.68_{\pm 0.23}$ & $53.02_{\pm 1.04}$ & $28.71_{\pm 0.16}$ & $51.12_{\pm 0.38}$ & $16.49_{\pm 0.34}$ & $32.76_{\pm 1.04}$ & $38.96_{\pm 0.15}$ \\
Image Features & ProbCover* & $75.34_{\pm 1.01}$ & $55.73_{\pm 0.49}$ & $34.36_{\pm 0.07}$ & $72.24_{\pm 1.34}$ & $16.55_{\pm 0.75}$ & $38.77_{\pm 2.59}$ & $48.83_{\pm 0.19}$ \\
Image Features & DropQuery & $77.97_{\pm 0.37}$ & $57.43_{\pm 0.33}$ & $37.39_{\pm 0.14}$ & $75.30_{\pm 0.48}$ & $17.09_{\pm 0.10}$ & $42.10_{\pm 0.34}$ & $51.21_{\pm 0.01}$ \\
\midrule
Class-guided Features & TypiClust & $81.08_{\pm 0.27}$ & $71.45_{\pm 1.57}$ & $40.87_{\pm 0.36}$ & $78.83_{\pm 0.14}$ & $18.10_{\pm 0.17}$ & $40.05_{\pm 0.74}$ & $\mathbf{55.06_{\pm 0.06}}$ \\
Class-guided Features & ProbCover & $63.95_{\pm 0.89}$ & $62.58_{\pm 1.91}$ & $32.61_{\pm 0.00}$ & $61.31_{\pm 0.32}$ & $15.81_{\pm 0.13}$ & $35.50_{\pm 0.51}$ & $\mathbf{46.99_{\pm 1.33}}$ \\
Class-guided Features & ProbCover* & $78.86_{\pm 1.06}$ & $68.61_{\pm 1.67}$ & $39.11_{\pm 0.17}$ & $77.07_{\pm 0.37}$ & $17.83_{\pm 0.62}$ & $38.40_{\pm 0.77}$ & $\mathbf{53.31_{\pm 0.28}}$ \\
Class-guided Features & DropQuery & $79.85_{\pm 0.32}$ & $70.95_{\pm 0.40}$ & $40.41_{\pm 0.53}$ & $77.86_{\pm 0.34}$ & $17.52_{\pm 0.36}$ & $39.11_{\pm 0.49}$ & $\mathbf{54.28_{\pm 0.10}}$ \\
\bottomrule
\end{tabular}
}
\end{table}

\vspace{1mm}
\noindent\textbf{Additional Clustering-based Baselines with Class-guided Features.}
In Table~\ref{tab:init_class_guided}, we first evaluate these baselines in the initial round using only image features and find that they mitigate the cold-start issue more effectively than Random. When class-guided features are added, all baselines benefit from a positive synergy that further enhances performance. Specifically, TypiClust~\citep{hacohen2023how}, ProbCover (K-means)~\citep{yehuda2022active}, and DropQuery~\citep{gupte2024revisiting} employ K-means clustering in the given feature space, selecting samples within each cluster based on typicality, coverage, or proximity to the centroid, respectively. Our method operates in the same manner as DropQuery~\citep{gupte2024revisiting} during initial pool selection, but differs in that we use class-guided features instead of image features. As shown in Table~\ref{tab:init_class_guided} and Figure~\ref{subfig:ablation-c}, the proposed class-guided features can be easily integrated into existing baselines and yield substantial performance gains.

\end{document}